\documentclass{article}

\usepackage[preprint]{neurips_2023}

\usepackage[utf8]{inputenc} 
\usepackage[T1]{fontenc}    
\usepackage{hyperref}       
\usepackage{url}            
\usepackage{booktabs}       
\usepackage{amsfonts}       
\usepackage{nicefrac}       
\usepackage{microtype}      
\usepackage{xcolor}         

\usepackage{multirow}
\usepackage{color}
\usepackage{colortbl}
\usepackage{stfloats}
\usepackage{comment}
\usepackage{proof}
\usepackage{fancybox}
\usepackage[flushleft]{threeparttable}
\usepackage{wrapfig}
\usepackage{pgfplotstable} 
\pgfplotsset{compat=1.12}
\usepackage{tabularx}
\usepackage{arydshln}
\usepackage{hyperref}
\usepackage[most]{tcolorbox}
\usepackage{subfig}
\usepackage[noend]{algpseudocode} 

\usepackage{bm}
\usepackage{mathtools}
\usepackage{stmaryrd}
\usepackage{graphicx}
\usepackage{color}
\usepackage{colortbl}
\usepackage{stfloats}
\usepackage{comment}
\usepackage{proof}
\usepackage{fancybox}
\usepackage[flushleft]{threeparttable}

\usepackage{paralist, tabularx}
\usepackage{wrapfig}

\usepackage{pgfplotstable} 
\pgfplotsset{compat=1.12}

\usepackage{tabularx}

\usepackage{arydshln}
\usepackage{hyperref}

\title{ISR-LLM: Iterative Self-Refined Large Language Model for Long-Horizon Sequential Task Planning}

\author{%
  Zhehua~Zhou \\\
  University of Alberta\\
  \texttt{zhehua1@ualberta.ca} \\
  \And
  Jiayang~Song \\\
  University of Alberta\\
  \texttt{jiayan13@ualberta.ca} \\
  \And
  Kunpeng~Yao \\\
  Swiss Federal Institute of Technology Lausanne (EPFL)\\
  \texttt{kunpeng.yao@epfl.ch} \\
  \And
  Zhan~Shu \\\
  University of Alberta\\
  \texttt{zshu1@ualberta.ca} \\
  \And
  Lei~Ma \\\
  The University of Tokyo \\
  University of Alberta\\
  \texttt{ma.lei@acm.org} \\
}

\begin{document}

\maketitle

\begin{abstract}
Motivated by the substantial achievements observed in Large Language Models (LLMs) in the field of natural language processing, recent research has commenced investigations into the application of LLMs for complex, long-horizon sequential task planning challenges in robotics.
LLMs are advantageous in offering the potential to enhance the generalizability as task-agnostic planners and facilitate flexible interaction between human instructors and planning systems.
However, task plans generated by LLMs often lack feasibility and correctness. 
To address this challenge, we introduce ISR-LLM, a novel framework that improves LLM-based planning through an iterative self-refinement process. 
The framework operates through three sequential steps: \emph{preprocessing}, \emph{planning}, and \emph{iterative self-refinement}. 
During preprocessing, an LLM translator is employed to convert natural language input into a Planning Domain Definition Language (PDDL) formulation. 
In the planning phase, an LLM planner formulates an initial plan, which is then assessed and refined in the iterative self-refinement step by using a validator. 
We examine the performance of ISR-LLM across three distinct planning domains.
The results show that ISR-LLM is able to achieve markedly higher success rates in task accomplishments compared to state-of-the-art LLM-based planners. 
Moreover, it also preserves the broad applicability and generalizability of working with natural language instructions. 
The code related to this work is available at \url{https://github.com/zhehuazhou/ISR-LLM}.
\end{abstract}

\section{Introduction}
\label{sec.intro}
Large Language Models (LLMs), underpinned by deep learning architectures, have recently revolutionized artificial intelligence (AI) by demonstrating unprecedented abilities in understanding, generating, and manipulating natural language text~\cite{bommasani2021opportunities, brown2020language, devlin2018bert, radford2019language, raffel2020exploring}.
This surge in LLM research has been accompanied by a growing interest in leveraging these models to tackle a diverse array of challenges across various research fields, including data analysis~\cite{agrawal2022large}, code generation~\cite{vaithilingam2022expectation}, reasoning~\cite{zelikman2022star}, robotic control~\cite{ahn2022can}, and so on.

Due to their rich internalized knowledge about the world~\cite{petroni2019language,davison2019commonsense}, LLMs have also garnered considerable attention within the field of long-horizon sequential task planning~\cite{roijers2013survey}.
Unlike short-term robotic planning problems, \emph{long-horizon sequential task planning} often involves devising interconnected actions that are spanned over extended timeframes to achieve control objectives.
Since the execution of actions at one point in time can greatly impact subsequent actions and outcomes, long-horizon planning is usually considered a more challenging problem due to its inherent intricacy in managing temporal dependencies and combinatorial complexity~\cite{hartmann2022long}, thereby necessitating innovative planning approaches that are able to balance the trade-offs between efficiency, optimality, and adaptability.

The traditional way to address long-horizon sequential task planning typically relies on first establishing a symbolic and logic-based representation of the planning problem~\cite{haslum2019introduction}, and then employing techniques such as state space search~\cite{zhang1999state} or heuristic search~\cite{edelkamp2011heuristic} to find a feasible solution.
However, this method usually requires the manual specification of symbolic planning domains, which demands a notable degree of expertise in the field.
Furthermore, many desirable properties of plans, e.g., user preferences, which can be specified in natural language by individuals without specialized training, may prove intricate or even infeasible to be encapsulated within formal logic frameworks. 
As a result, the adaptability of conventional methods is constrained, limiting their utility in diverse contexts.

To overcome this limitation, there is a growing trend in recent studies to explore the potential of utilizing LLMs as task-agnostic reasoning modules, with the aim of facilitating more generalized and intelligent robotic planning~\cite{ahn2022can,huang2022inner}.
Leveraging their pre-trained knowledge, these LLM-based planners are able to effectively comprehend both explicit human-generated natural language directives and the inherent constraints interwoven within planning tasks~\cite{huang2022language}. 
This greatly reduces the necessity for labor-intensive manual rule encoding and circumvents the need for intricate specification of symbolic planning domains~\cite{lin2023text2motion}. 
Moreover, the intuitive nature of textual prompts allows for seamless interactions between LLM-based planners and human instructors, facilitating the integration of human expertise into the planning process.
However, the efficacy and reliability of such LLM-based planners are often not satisfying due to the inherent design and training methodologies of LLMs.
LLMs are essentially engineered to generate word sequences that align with human-like context, yet the assurance of their planning capabilities is not guaranteed~\cite{brown2020language}.
Recent investigations have revealed instances where the correctness of generated actions and the success rate of task accomplishment by LLM-based planners fall short of expectations~\cite{valmeekam2022large}. 
This limitation becomes further pronounced in long-horizon sequential task planning, where complex action dependencies and extended temporal considerations introduce additional difficulties that challenge the planning abilities of LLMs.

\begin{figure*}
    \centering
    \includegraphics[width=\linewidth]{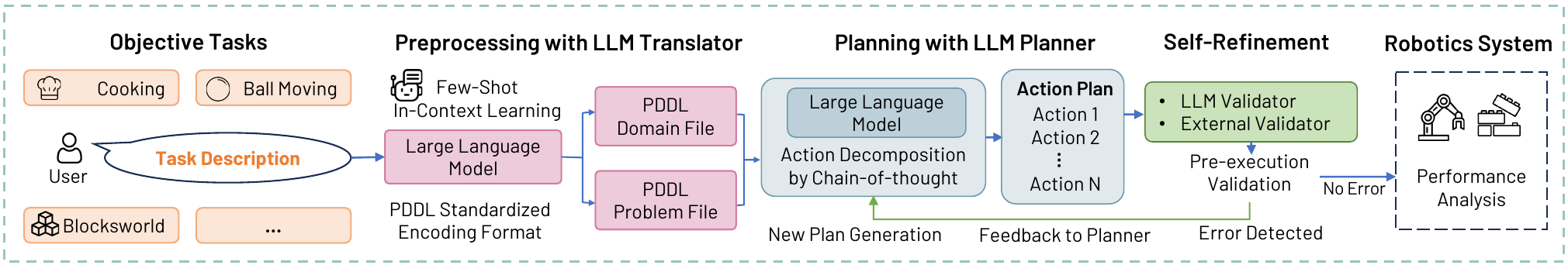}
    \caption{Overview of the proposed ISR-LLM framework. 
    It consists of three steps: preprocessing, planning, and iterative self-refinement.}
    \label{fig.overview}
\end{figure*}

In this work, we aim to enhance the performance of LLM in long-horizon sequential task planning. 
Drawing inspiration from recent research that reveals the potential for LLM improvements through self-refinement~\cite{madaan2023self, huang2022large}, we propose the Iterative Self-Refined LLM (ISR-LLM) framework that utilizes the power of iterative self-refinement to improve planning outcomes.
Our framework consists of three steps (see Fig.~\ref{fig.overview}): 
(1) \textit{Preprocessing}, where an LLM translator is employed to translate the natural language inputs into their respective Planning Domain Definition Language (PDDL)~\cite{haslum2019introduction} formulations;
(2) \textit{Planning}, where an LLM planner takes the translated PDDL problem as input and determines the action sequence to accomplish the long-horizon sequential task planning;
(3) \textit{Iterative self-refinement}, where a validator is used to examine the correctness of the generated action plan and provide feedback to the LLM planner. 
Then based on the feedback, the LLM planner performs the iterative self-refinement process to find a revised action plan.
We consider two different types of validators in our approach: an LLM-based self-validator and an external validator that leverages auxiliary verification tools.

Through comprehensive experiments across diverse planning problem domains, we show that, compared to state-of-the-art approaches, ISR-LLM achieves better feasibility and success rate in long-horizon sequential task planning.
The contributions of this work are threefold:
\begin{itemize}
    \item We present ISR-LLM, a novel framework achieved by integrating a self-refinement mechanism into LLM. 
    This approach addresses long-horizon sequential task planning and offers remarkable advancements in both feasibility and correctness.
    
    \item We introduce and evaluate the effectiveness of two types of validators, i.e., an LLM-based self-validator and an external validator, in providing feedback to the LLM planner for executing the iterative self-refinement process.
    
    \item We highlight the superiority of our proposed framework in comparison to contemporary state-of-the-art methods, through an investigation of ISR-LLM across three diverse planning domains.
    
\end{itemize}

\section{Related Work}
\label{sec.related_work}

\subsection{Long-Horizon Sequential Task Planning}
\label{sec.classical_planning}

Long-horizon sequential task planning aims to find an optimal action sequence capable of accomplishing a specified task objective~\cite{helmert2006fast}.
In recent robotic studies, PDDL or Answer Set Programming (ASP)~\cite{brewka2011answer} are often utilized as the language for representing the planning problems~\cite{jiang2019task}.
A prevalent method employed to tackle these planning tasks is to utilize a search-based or sampling-based algorithm to find a viable plan~\cite{levine2003learning,segovia2021generalized,cohen2010search}.
This strategy has found successful applications across diverse robotic domains, e.g., mobile robots~\cite{zhang2015mobile}, autonomous vehicles~\cite{ding2020task}, and robotic manipulators~\cite{garrett2020pddlstream}.
However, these approaches rely on a predetermined symbolic and logical representation of the planning domain, which usually demands a high level of expert knowledge for formulation. 
Moreover, due to the inherent abundance of potential action options associated with long-horizon sequential task planning, search-based or sampling-based strategies may encounter impediments in such scenarios.
Some approaches also use example plans to construct novel plans, which are often represented through a finite state machine~\cite{levesque2005planning, winner2008learning}.
However, finding a useful example plan may be challenging or even impossible within certain task scenarios.

It is also worth mentioning that, another important category of robotic planning is Task and Motion Planning (TAMP)~\cite{garrett2021integrated}, which combines high-level task planning in discrete spaces and low-level robot motion planning in continuous space as a hierarchical planning framework.
In TAMP, the focus extends beyond mere task planning to encompass the executability of the determined actions, i.e., the actions must be executable by the robot with a viable motion trajectory that is subject to both robotic and environmental constraints~\cite{toussaint2015logic, driess2019hierarchical}.
However, how to accurately ground actions generated by LLMs into feasible robot motions remains a challenging and ongoing area of research~\cite{ahn2022can, huang2022inner}.
Therefore, in this work, we focus only on exploring the task planning capabilities of LLMs. 

\subsection{Planning with LLM}
\label{sec.planning_LLM}

To overcome the limited generalizability of traditional task planners, researchers have started investigating the possibility of utilizing LLMs as task-agnostic planners~\cite{sharma2021skill,li2022pre,zeng2022socratic,singh2023progprompt}. 
A multitude of studies have delved into grounding the language commands generated by LLMs to executable robotic actions~\cite{ahn2022can,huang2022inner,ding2023task,lin2023text2motion}.
For instance, in~\cite{ahn2022can}, scores are assigned to potential actions through a value function, and the action with the highest likelihood of success is selected. 
Similarly, \cite{huang2022language} adopts prompt engineering to extract actions that are executable for the robots. 
In~\cite{huang2022inner}, environmental feedback is introduced to enable online adjustment of action plans that are infeasible for the robots.
Although the focus of this work is not the grounding of actions, these studies illustrate the competencies of LLMs in addressing diverse robotic planning tasks.

Besides grounding language instructions, recent studies have also sought to combine LLMs with PDDL as a means of elevating the performance of LLM-based planners~\cite{valmeekam2022large,silver2022pddl,silver2023generalized,liu2023llm}.
In~\cite{valmeekam2022large}, a Blocksworld~\cite{slaney2001blocks} benchmark is proposed to assess the LLM's capability in handling natural language inputs for planning. 
However, the results reveal a discouraging performance of LLMs in long-horizon task planning, even within seemingly uncomplicated tasks.
In~\cite{silver2022pddl,silver2023generalized}, instead of natural language inputs, planning problems in PDDL syntax are directly presented to LLMs for generating action sequences. 
While this strategy contributes to enhanced performance, it inevitably diminishes the LLM's generalizability and often demands additional effort and expert knowledge for composing the corresponding PDDL files. 
In~\cite{liu2023llm}, LLM is employed not as a planner, but rather as a translator that converts natural language inputs into PDDL problems, which are subsequently solved using classical PDDL planners. 
However, such an approach requires an external solver, potentially impeding the wider applicability of LLMs as task-agnostic planners.
An analogous notion akin to our self-refinement concept is introduced in~\cite{raman2022planning}. 
After the generation of an action plan based on natural language inputs, it collects the error information returned from the execution of the plan. 
This information is then constructed as re-prompts that direct the LLM towards correcting the erroneous actions.
However, such a refinement process occurs subsequent to the action execution phase.
Our approach, in comparison, not only considers the utilization of an external validator to perform a similar self-refinement process, but also investigates the potential of LLMs for enabling pre-execution action corrections through self-validation capabilities.

\section{Preliminary}
\label{sec.preliminary}

\subsection{Task Planning}
\label{sec.planning_problem}

In this work, we consider the problem of task planning in a setting with discrete and fully observable states, finite actions, and deterministic transitions.
Such a problem $P$ is often represented by a tuple $P = \left<S, A, T, s_{\mathrm{init}}, G\right>$.
For each state $s \in S$ within the discrete set of states $S$, an action $a \in A$ can be selected from the set of applicable actions $A(s) \subseteq A$, i.e., the preconditions of the action $a$ must be fulfilled. 
The transition function $T: S \times A \rightarrow S$ determines the next state based on the current state and the selected action.
$s_{\mathrm{init}} \in S$ represents the initial state and $G \subseteq S$ is a set of goal states.
A solution to the planning problem $P$ is a sequential action plan $\pi = (a_1, a_2, \ldots, a_{n})$ that controls the initial state $s_{\mathrm{init}}$ to a goal state, i.e., we have $s_{i+1} = T(s_i, a_i)$ satisfied for all $0 \leq i \leq n$ and $s_{n+1} \in G$. 
For long-horizon sequential task planning, the number of actions $n$ tends to be relatively large.
In this work, we focus on investigating the capabilities of LLM in solving the designated task planning problem $P$.
Thus, our primary focus is the feasibility and success rate of planning rather than its optimality.

\subsection{PDDL}
\label{sec.pddl}

PDDL is a standardized encoding format designed for classical planning problems~\cite{aeronautiques1998pddl,fox2003pddl2}.
A planning problem $P$ represented in PDDL syntax consists of two files: a domain file and a problem file.
The domain file embodies the foundational rules of the planning domain. 
It not only defines the predicates that elucidate the configuration of the state space $S$, but also formulates the preconditions and effects of all possible actions $a\in A$, i.e., the transition function $T$.
The problem file is used to define the available objects within the planning domain, as well as the initial state and goal conditions. 
Concrete examples of PDDL domain and problem files for the experiments considered in this work can be found in Appendix~\ref{sec.translator_example}.
In this work, we assume that the natural language input provided to the LLM should include both the initial state and the goal conditions, such that the LLM translator is able to convert it into corresponding PDDL files.
For more details about PDDL, we direct the interested readers to~\cite{haslum2019introduction}.

\section{ISR-LLM}
\label{sec.isr-llm}

In this section, we introduce ISR-LLM, a novel framework that utilizes iterative self-refinement to find an action plan with improved accuracy and feasibility.
It includes three steps: preprocessing with an LLM translator, planning with an LLM planner, and iterative self-refinement loop with a validator that is selected from either an LLM-based self-validator or an external validator. 
Details are explained as follows.

\subsection{Preprocessing with LLM Translator}
\label{sec.translator}

As illustrated in Fig.~\ref{fig.overview}, the LLM translator first converts the given natural language instructions into a PDDL formulation, specifically representing them using the domain and problem files.
The rationale for employing such a translator is grounded in its notable advantages, even though an LLM planner could be designed to operate directly on natural language inputs, as demonstrated in~\cite{lin2023text2motion}.
The adoption of a formal representation, i.e., PDDL, offers twofold benefits to the subsequent validation process of the generated plan.
Firstly, it enables the usage of existing PDDL validators as the external validator, e.g., VAL~\cite{howey2004val} or PDDL.lj~\cite{zhi2022pddl}.
This obviates the necessity of developing a custom validator and thereby saves substantial time and effort.
Secondly, rather than relying solely on language cues, this approach enables the LLM-based self-validator to acquire a comprehension akin to a state-machine understanding of the system state. 
This, in turn, facilitates a more precise evaluation of the correctness of the selected actions.

In order to ensure the structural accuracy of the translated PDDL files, we adopt a technique known as few-shot in-context learning~\cite{brown2020language}. 
This technique involves embedding illustrative examples within the prompt, effectively instructing the LLM on how to formulate responses to given queries in a desired manner. 
Similar to~\cite{liu2023llm}, we assume that the domain-specific knowledge pertinent to each considered planning task is available in advance, and thus include it within the few-shot examples provided to the LLM translator.
An example of the prompt presented to the LLM translator for the Blocksworld planning domain (see Sec.~\ref{sec.experimental_setup} for a detailed explanation about this domain) is shown in Fig.~\ref{fig.prompt_examples}, and a complete list of all employed few-shot examples within this work is given in Appendix~\ref{sec.translator_example}.

\begin{figure*}
    \centering
    \includegraphics[width=\linewidth]{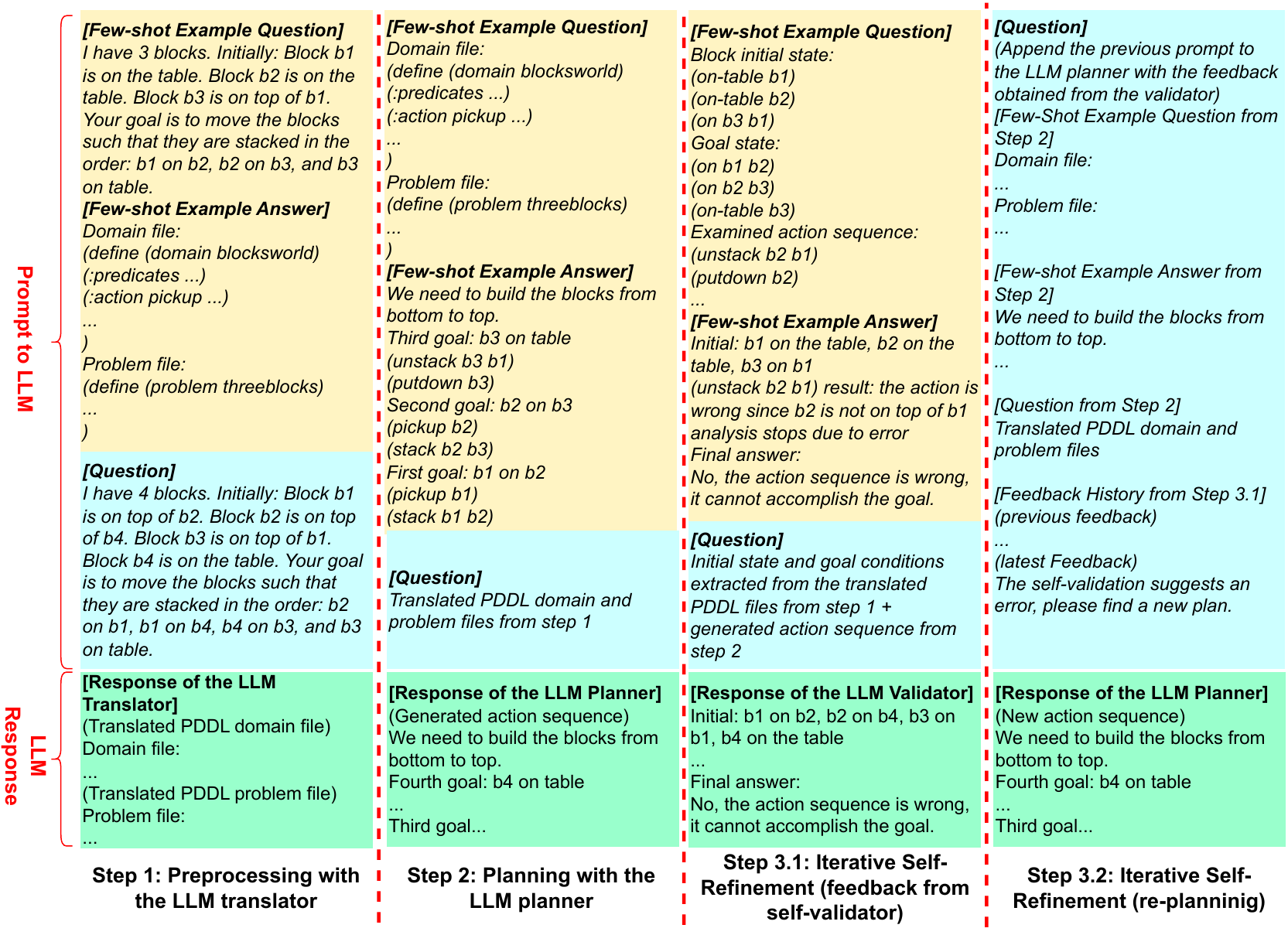}
    \caption{Examples of the prompts used in ISR-LLM. 
    The prompt provided to the LLM contains two parts: the few-shot examples (shaded with a yellow color) and the actual question (blue). 
    Details about the few-shot examples are given in Appendix~\ref{sec.appendix_prompt}.
    The texts shaded with a green color represent the LLM's responses. 
    The LLM translator first converts the natural language instructions into PDDL domain and problem files.
    Then, an initial plan is generated using the translated files, which is subsequently revised through an iterative self-refinement process.
    }
    \label{fig.prompt_examples}
\end{figure*}

\subsection{Planning with LLM Planner}
\label{sec.planner}

Once the natural language input is translated, the LLM planner takes these PDDL files as inputs and determines an action sequence aimed at achieving the given task (see Fig.~\ref{fig.overview}). 
In addition to few-shot in-context learning, we also integrate the Chain-of-Thought (CoT) technique~\cite{wei2022chain} into the prompts provided to the LLM planner.
CoT operates by decomposing the overall problem into intermediate steps, thus enabling the LLM to tackle complex reasoning problems that may not be solvable via standard prompting methods. 
An illustrative example of the prompt presented to the LLM planner is given in Fig.~\ref{fig.prompt_examples}, and a comprehensive list of all the employed few-shot examples is accessible in Appendix~\ref{sec.planner_example}.

Within this step, we obtain an initial action plan for addressing the given planning problem. 
Subsequently, as detailed in the next subsection, such an initial plan is examined by a validator. 
Utilizing the feedback received from the validator, the LLM planner performs a self-refinement to find a new plan that attempts to correct erroneous actions.

\subsection{Iterative Self-Refinement Loop with Validator}
\label{sec.validator}

The central component of the iterative self-refinement loop is the validator, as demonstrated in Fig.~\ref{fig.overview}.
Through the examination of the generated action sequence, the validator constructs feedback, pinpointing any actions considered incorrect, and subsequently conveys this information to the LLM planner.
Then based on the feedback, the LLM planner initiates a self-refinement process to rectify the incorrect action and devise a new action plan.
Note that, while the generated action sequence may contain multiple errors, analyzing actions subsequent to the initial error is often unnecessary, since the first error could potentially render the foundation of all ensuing actions fundamentally flawed.
Thus, the self-refinement process is executed iteratively within a loop, where in each step, the validator stops at the first identified error. 
The information concerning this error is then returned, ensuring that each iterative stage is solely focused on rectifying this detected mistake. 
The iterative self-refinement loop persists until either the validator identifies no errors or a predefined maximum number of iterations is reached. 
The action sequence, resulting from the iterative self-refinement loop, is then accepted as the final generated action sequence.

We consider two types of validators: a self-validator, which employs the LLM to assess the correctness of the generated action plan, and an external validator, which leverages external tools for performing the analysis.
It is worth mentioning that, although the external validator is capable of providing accurate feedback on the feasibility of the generated plan, its implementation often demands a considerable amount of effort and may be unavailable for certain tasks. 
Conversely, the usage of an LLM as an internal self-validator economizes both time and effort.
However, it has the inherent risk of possibly yielding imprecise or even erroneous feedback.
The selection of the validator type, therefore, hinges upon the specific evaluation requirements and the context of the validation scenario.

An example of the prompts provided to the LLM-based self-validator is shown in Fig.~\ref{fig.prompt_examples}, where few-shot learning and CoT techniques are also employed.
All examples used for the experimental domains explored in this work are given in Appendix~\ref{sec.validator_example}.

\section{Experimental Results}
\label{sec.results}
To evaluate the performance of ISR-LLM in long-horizon sequential task planning, we perform experiments across three diverse planning domains. 
Moreover, we also investigate the influence of different LLMs on the performance of ISR-LLM, as well as the impact of the LLM translator. 
A detailed explanation of the experimental setup and results is provided in the following subsections.

\subsection{Experimental Setup}
\label{sec.experimental_setup}

\begin{figure*}[!t]
\centering
\subfloat[Cooking]{\includegraphics[height = 2in]{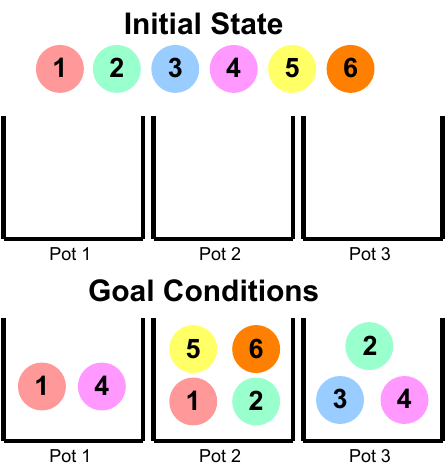}%
\label{fig.cooking}}
\hfil
\subfloat[Blocksworld]{\includegraphics[height = 2in]{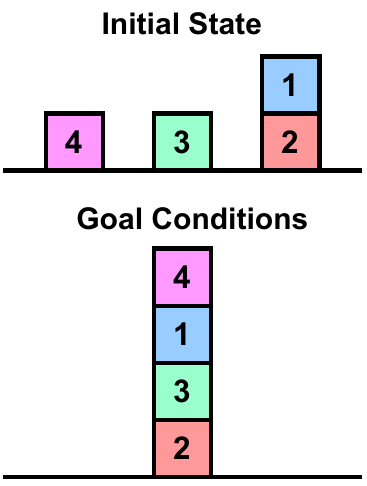}%
\label{fig.blocksworld}}
\hfil
\subfloat[Ball Moving]{\includegraphics[height = 1.6in]{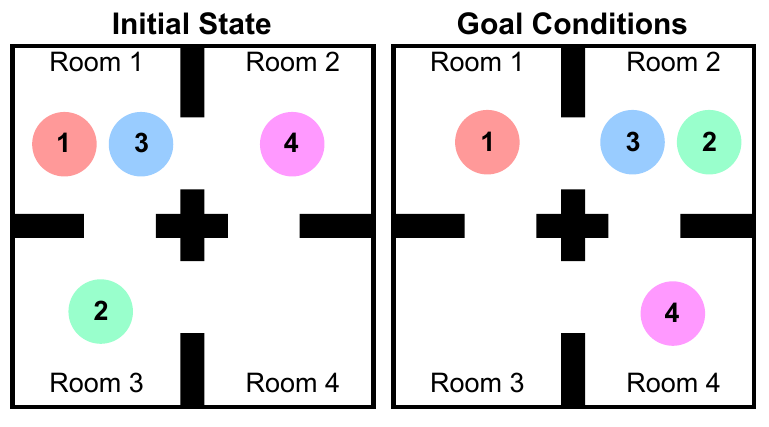}%
\label{fig.ballmoving}}
\hfil
\caption{Three planning domains used in this work.}
\label{fig.experimental_setup}
\end{figure*}

We utilize the following three planning domains as benchmark problems to evaluate the performance of ISR-LLM. 
These domains are derived from existing literature and are extensively employed in planning research~\cite{liu2023llm, silver2023generalized,valmeekam2022large,silver2022pddl}. 
Detailed examples about each planning domain are presented in Appendix~\ref{sec.appendix_prompt}.

\begin{itemize}
    \item \textit{Cooking:} There are $n$ pots and a total of 6 different ingredients (see Fig.~\ref{fig.cooking}). 
    The robot's task is to add ingredients to each pot according to a prescribed recipe. 
    Each pot possesses its own randomly generated recipe, which stipulates the inclusion of 2 to 4 different ingredients. 
    The robot has three actions: picking up an ingredient, putting down an ingredient, and adding the ingredient to a pot. 
    A constraint that must be fulfilled is that each ingredient may only be retrieved once by the robot, i.e., once the robot has picked up an ingredient, it must distribute it to all pots that require this ingredient as per their individual recipes. 

    \item \textit{Blocksworld:} There are $n$ blocks, initially randomly placed on a table. 
    The objective of the robot is to assemble these blocks into a stack, adhering to a specific prescribed order (see Fig.~\ref{fig.blocksworld}). 
    The robot has four actions: picking up a block that is on the table, putting down a block that is currently in its hand onto the table, unstacking a block from the top of another block to hold it in its hand, and stacking the block that is currently in its hand on top of another block. 
    However, the robot can only manipulate one block at a time, i.e., any block that has other blocks situated on top of it is considered fixed.
    
    \item \textit{Ball Moving:} There are $n$ balls, initially randomly distributed among 4 rooms (see Fig.~\ref{fig.ballmoving}). 
    The robot needs to relocate the balls to their predefined goal rooms, with the constraint that it can hold no more than one ball at a time. 
    The robot has three actions: picking up a ball, putting down a ball, and moving from its current room to another room.
    
\end{itemize}

For all three planning domains, we investigate two specific cases with $n=3$ and $n=4$, to examine the influence of the number of objects, which is directly correlated with the complexity of the task, on the performance of the proposed ISR-LLM framework.
Furthermore, to evaluate the impacts of various LLMs on the planning outcomes, we employ two LLMs, namely GPT3.5 and GPT4, and compare their capabilities in task planning within the ISR-LLM framework.

For each planning task, we evaluate three different methods:
(1) \textit{LLM-direct}, which is the baseline approach grounded in~\cite{silver2023generalized,silver2022pddl,valmeekam2022large}. 
It leverages the LLM to formulate an action plan directly from the given PDDL input. 
To ensure a fair comparison with ISR-LLM, we utilize the LLM translator to convert natural language inputs into PDDL files in this method.
(2) \textit{ISR-LLM-self}, which employs the ISR-LLM framework with an LLM-based self-validator;
(3) \textit{ISR-LLM-external}, which incorporates an external validator to generate feedback for ISR-LLM. 
In order to mitigate the influence of existing PDDL validators and focus on analyzing the performance of ISR-LLM, we implement our own custom external validators in this work.

We randomly generate 30 unique cases with varying initial states and goal conditions for each planning task. 
The few-show examples used for the LLM translator, the LLM planner, and the LLM-based self-validator are given in Appendix~\ref{sec.appendix_prompt}.
All LLM's responses during the experiments are presented in our website\footnote{\url{https://github.com/zhehuazhou/ISR-LLM}}.
The success rates of task accomplishments for the three aforementioned methods are recorded. 
All experiments are conducted on a laptop equipped with an Intel(R) Core(TM) i7-10870H CPU @ 2.20GHz Processor with 8 CPUs, and an NVIDIA RTX 3080 Max-Q GPU with 16 GB VRAM.
The detailed results are presented in the next subsection.

\subsection{Performance of ISR-LLM}
\label{sec.performance_results}

\begin{table*}[t]
\centering
\caption{Success rate of ISR-LLM in different planning domains.}
\label{table.results}
\resizebox{1\textwidth}{!}{
\begin{tabular}{c|ccc|ccc}
\toprule
& \multicolumn{3}{c|}{GPT3.5} & \multicolumn{3}{c}{GPT4}      \\
Planning domain & LLM-direct & ISR-LLM-self & ISR-LLM-external & LLM-direct & ISR-LLM-self & ISR-LLM-external \\ \midrule
Cooking ($n=3$) & 47\% & 67\% &  100\% &  100\% &  100\% & 100\% \\  \hline
Cooking ($n=4$) & 40\% & 53\% &  63\% &  100\% &  100\% &  100\% \\  \hline
Blocksworld ($n=3$) & 20\% & 37\% &  70\% &  43\% &  60\% & 97\% \\  \hline
Blocksworld ($n=4$) & 10\% & 17\% &  53\% &  40\% &  60\% & 80\% \\  \hline
Ball Moving ($n=3$) & 33\% & 50\% & 70\% &  93\% &  100\% & 100\% \\  \hline
Ball Moving ($n=4$) & 17\% & 27\% &  57\% &  90\% &  93\% & 97\% \\  \hline
\end{tabular}
}
\end{table*}

The results of the experiments are summarized in Table~\ref{table.results}. 
In the cases utilizing GPT3.5, the proposed ISR-LLM framework demonstrates a notable enhancement in success rates across all planning domains when compared to the baseline approach.
While the LLM-based self-validator contributes to an approximate $15\%$ increase in performance, the external validator can further amplify the success rate by roughly $40\%$ to $50\%$.
The only exception occurs in the case $n=4$ for the Cooking domain, where a $23\%$ increase is observed.
This might be attributed to the excessive number of required actions in this planning task, rendering LLMs less effective at correcting errors.

The success rates are also influenced by task complexity, as indicated by the number of objects. 
Increases in object numbers correspond to decreased success rates in the Cooking, Blocksworld, and Ball Moving domains for all three approaches (LLM-direct: $-7\%$, $-10\%$, $-16\%$; ISR-LLM-self: $-14\%$, $-20\%$, $-23\%$; ISR-LLM-external:$-37\%$, $-17\%$, $-13\%$).
This trend reflects the increased difficulty in rectifying erroneous actions as the planning horizon extends.
Moreover, the success rate varies among planning domains. 
Compared to the Cooking and the Ball Moving domains, the Blocksworld domain, which demands more sophisticated logical thinking, demonstrates lower success rates.
Nevertheless, the proposed ISR-LLM is still able to improve the planning outcomes within this domain.

It can also be observed that GPT4 greatly outperforms GPT3.5 in long-horizon sequential task planning, corroborating the common assertion that GPT4 possesses a markedly superior reasoning capability. 
The baseline method, i.e., LLM-direct, when coupled with GPT4, is able to achieve a success rate exceeding $90\%$ in the Cooking and the Ball Moving domains, where ISR-LLM also maintains this high performance level. 
However, in the more logically complex Blocksworld domain, GPT4 demonstrates diminished performance using the baseline approach.
Nevertheless, the employment of ISR-LLM also elevates the success rate for this domain, with the self-validator contributing an increase of about $20\%$, and the external validator enhancing it by more than $40\%$.
Interestingly, the influence of the number of objects appears to be less pronounced when GPT4 is utilized.
This may be attributed to GPT4's enhanced reasoning capabilities, which facilitate more effective logical thinking, and thereby mitigate the impact of the number of objects on the results.


\subsection{Influence of the LLM Translator}
\label{sec.results_translator}

\begin{table}[t]
\centering
\caption{Success rate of ISR-LLM with and without the LLM translator in Blocksworld domain with $n=3$ and GPT3.5.}
\label{table.translator}
\begin{tabular}{c|c|c}
\toprule
Method & With LLM Translator & Without LLM Translator \\ \hline
LLM-direct &  20\% & 13\% \\ \hline 
ISR-LLM-self & 36\%  & 16\% \\ \hline
ISR-LLM-external & 70\% & 63\% \\ \hline
\end{tabular}
\end{table}

We also evaluate the influence of the LLM translator using the Blocksworld domain with $n=3$ and GPT3.5 as an example, as this case demonstrates where the efficacy of ISR-LLM is most obvious. 
By omitting the LLM translator and directly utilizing natural language input, we compare the success rates of task planning and present the results in Table~\ref{table.translator}.
It can be observed that, while the LLM translator slightly improves the planning performance of the baseline approach, the self-validator greatly benefits from the translator, showing a $20\%$ increase in the success rate. 
The reason could be that the translated PDDL files offer a symbolic and logical representation of the planning domain, thereby allowing the LLM to form a more concrete understanding of the system state, as opposed to relying solely on linguistic cues. 
In contrast, the performance of the external validator remains relatively consistent, irrespective of the presence of the LLM translator.
This consistency arises from our custom validator's ability to provide accurate feedback, whether PDDL formulations are employed or not.
However, as previously mentioned, introducing translated PDDL files enables the usage of existing PDDL validators, potentially saving substantial time and effort needed for implementing a custom validator.

\subsection{Grounding the Actions}
\label{sec.results_grounding}

\begin{figure*}[!t]
\centering
\subfloat[Unstack b1 from b2]{\includegraphics[width= 1.3in]{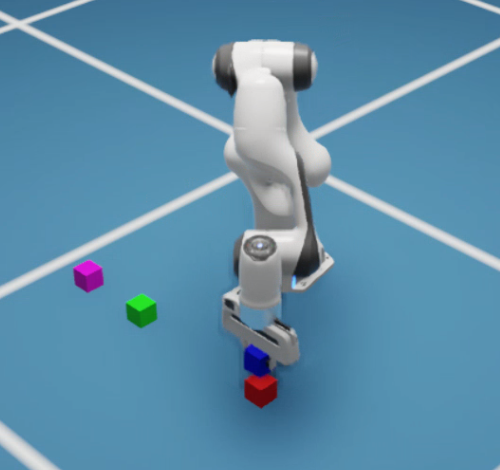}
\label{fig.step1}}
\hfil
\subfloat[Put down b1]{\includegraphics[width= 1.3in]{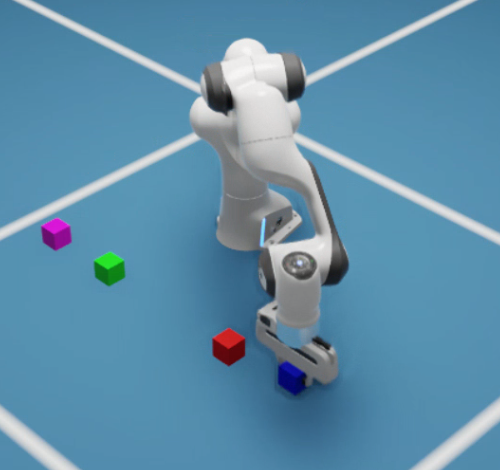}
\label{fig.step2}}
\hfil
\subfloat[Pick up b3]{\includegraphics[width= 1.3in]{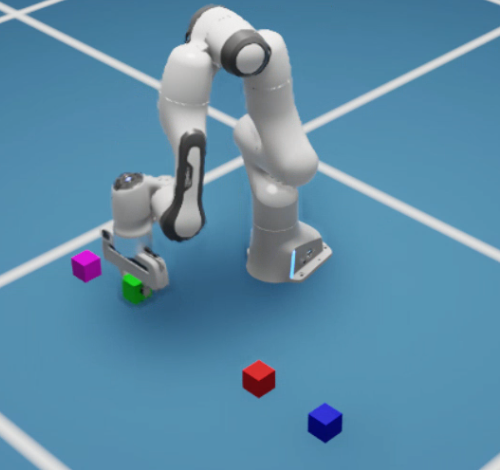}
\label{fig.step3}}
\hfil
\subfloat[Stack b3 on b2]{\includegraphics[width= 1.3in]{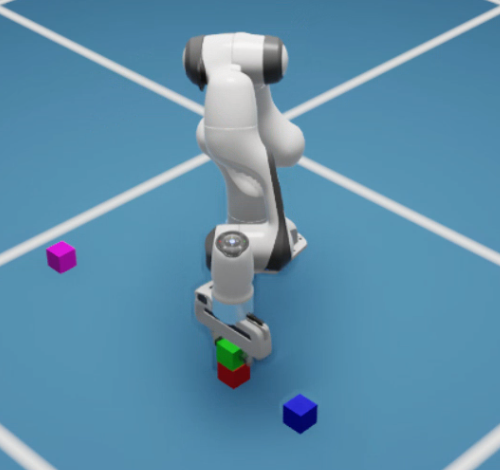}
\label{fig.step4}}
\hfil
\subfloat[Pick up b1]{\includegraphics[width= 1.3in]{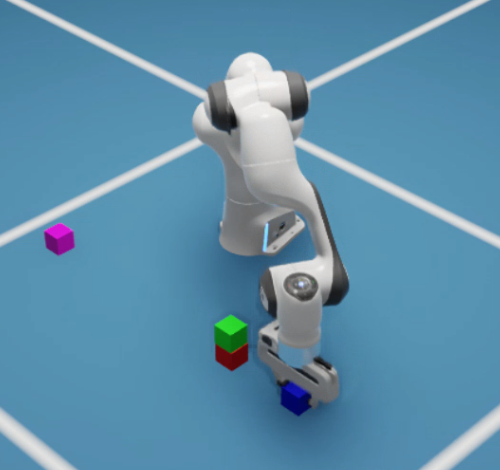}
\label{fig.step5}}
\hfil
\subfloat[Stack b1 on b3]{\includegraphics[width= 1.3in]{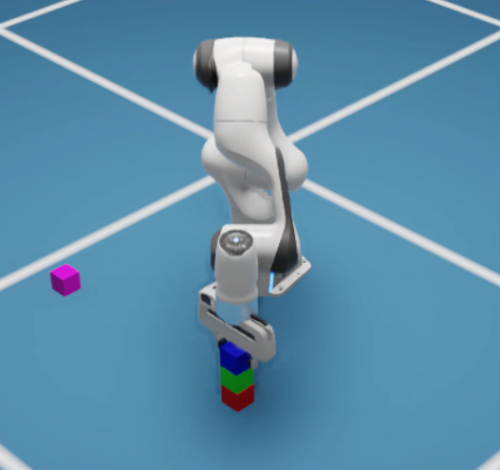}
\label{fig.step6}}
\hfil
\subfloat[Pick up b4]{\includegraphics[width= 1.3in]{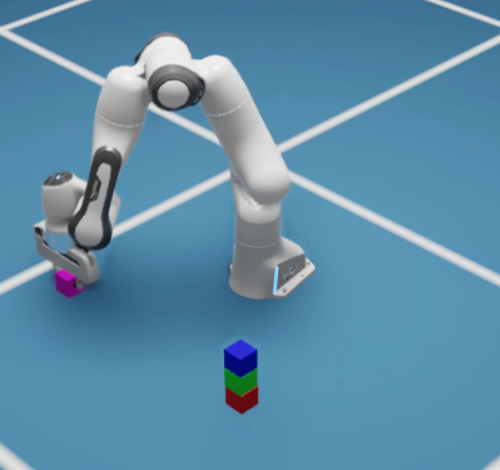}
\label{fig.step7}}
\hfil
\subfloat[Stack b4 on b1]{\includegraphics[width= 1.3in]{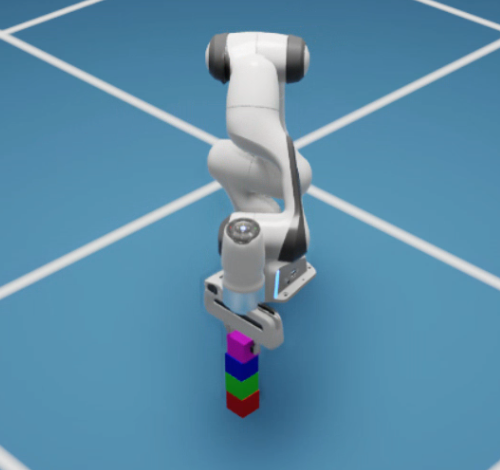}
\label{fig.step8}}
\hfil
\caption{Grounding of actions in the Blocksworld domain with four blocks. Initially, block b2 (red), b3 (green), b4 (pink) are on the table, and block b1 (blue) is on top of block b2. 
The goal is to stack the blocks in the given order: b4 on b1, b1 on b3, b3 on b2, and b2 on the table. }
\label{fig.ground_action}
\end{figure*}

Although it is beyond the scope of this work, we further demonstrate that the generated action plan can be directly grounded into feasible robot actions when paired with a suitable motion planner.
This highlights another advantage of employing the LLM translator within the ISR-LLM framework, as the use of PDDL formulation ensures that each generated action conforms to a predefined definition and structure.
Consequently, this simplifies the task of the motion planner in converting the action plan into executable robot movements. 
Figure~\ref{fig.ground_action} illustrates this grounding process, using an example from the Blocksworld domain with four blocks. 
Here, a pick-and-place controller is employed to execute the four different types of actions, assuming the robot knows the locations of the blocks.
The simulation is conducted in NVIDIA Omniverse Isaac Sim\footnote{\url{https://developer.nvidia.com/isaac-sim}}.

\section{Discussion}
\label{sec.discussion}

\paragraph{Self-Validator and External Validator}
Generally, the external validator is capable of providing feedback to a degree of precision that identifies the exact action in which an error resides.
Conversely, the self-validator usually only provides an overarching estimation regarding the correctness of the entire generated action plan. 
As a consequence, the external validator often leads to superior performance, as precise feedback can greatly facilitate the correction of erroneous actions. 
This benefit becomes more obvious as the planning horizon extends, or when complex logical thinking is demanded. 
However, as aforementioned, the external validator requires additional design and implementation effort. 
In contrast, the self-validator is advantageous in that it can be easily and directly employed without necessitating extra work.
Therefore, the selection between these validator types should be carefully considered in light of the specific task requirements and the resources available.

\paragraph{Planning Domains}
The planning capabilities of LLMs are influenced by the inherent characteristics of the planning domains. 
As observed from our experimental results, LLMs appear to excel in planning tasks that focus on adhering to specific instructions, such as Cooking, or performing repeated actions with identifiable patterns, e.g., Ball Moving. 
Conversely, when the planning tasks demand more complex logical thinking, as seen in the Blocksworld domain, their planning performance tends to diminish.
This phenomenon is more pronounced in the GPT4 cases. 
The underlying reason could be that LLMs are essentially trained to generate word sequences that mirror human-like thought processes, which suits tasks requiring instruction or pattern following. 
However, when critical logical reasoning becomes a vital component of the task, the inherent reasoning abilities of the LLMs become more important.
This suggests that enhancing the reasoning capabilities of LLMs could be a priority when aiming to utilize them as planners for more intricate planning tasks.

\paragraph{Limitations}
One limitation of the current LLM-based planners - even with the proposed ISR-LLM framework - is that the overall success rate often fails to exceed that of traditional search-based planners. 
However, as an initial exploratory work, we demonstrate the potential of utilizing LLM as a versatile and task-agnostic planner. 
This could significantly facilitate the deployment of various robotic systems across diverse scenarios and minimize the required effort in planning system design.
Moreover, the planning abilities of the ISR-LLM framework may see substantial improvements through refinements in the underlying reasoning capabilities of the LLMs. 
This could be potentially achieved through parameter fine-tuning technologies, such as integrating a fine-tuned LLM specifically designed for task planning.
Another limitation stems from the inherent randomness within LLMs, complicating assurances such as correctness or constraint satisfaction in the generated action plan. 
Therefore, the employment of LLMs may be inappropriate for certain tasks, especially those that are safety-critical.

\section{Conclusion}
\label{sec.conclusion}

In this paper, we explore the potential of leveraging LLMs for long-horizon sequential task planning based on natural language input.
To improve the correctness of the generated action plan, we introduce the ISR-LLM framework, which employs an iterative self-refinement approach for automatic plan revisions.
This framework consists of three steps.
First, an LLM translator converts the natural language input into a PDDL formulation, represented by PDDL files.
Second, using these translated PDDL files, an LLM planner formulates an initial action plan.
Third, an iterative self-refinement loop is initiated, wherein either an LLM-based self-validator or an external validator provides feedback on the correctness of the action plan, allowing the LLM planner to make necessary revisions to the action plan.
Through extensive experiments across three diverse planning domains, we demonstrate that ISR-LLM surpasses the performance of existing state-of-the-art LLM-based planners in long-horizon sequential task planning.
While maintaining the flexibility and generalizability to work with natural language input, our ISR-LLM framework consistently achieves high success rates in task accomplishments.
For future work, we plan to incorporate motion planning within the ISR-LLM framework, aiming to facilitate reliable and efficient task and motion planning across various robotic application scenarios.

\bibliographystyle{ACM-Reference-Format}
\bibliography{ref}

\newpage
\appendix
\section{Prompt Few-Shot Examples}
\label{sec.appendix_prompt}

\subsection{Translator}
\label{sec.translator_example}

\paragraph{Cooking:} 
For the Cooking planning domain, we use a total of three few-shot examples, which are given as follows.

\begin{tcolorbox}[title={Example 1},breakable]
\textbf{Question:}
I have 3 pots and 6 different ingredients. Each ingredient can only be picked up once. Your goal is to add ingredients to pots by following the receipts: pot1 contains ingredient1, ingredient4. pot2 contains ingredient2, ingredient3, ingredient5, ingredient6. pot3 contains ingredient1, ingredient4, ingredient5.

\textbf{Answer:}

Domain file: \\
\texttt{
(define (domain cooking) \\
(:types pot ingredient) \\
(:predicates  \\
(arm-empty) \\
(pot-empty ?p - pot) \\
(holding ?i - ingredient) \\
(contain ?p - pot ?i - ingredient) \\ 
) \\
(:action pick \\ 
  :parameters (?i - ingredient) \\
  :precondition (and (arm-empty)) \\
  :effect (and (holding ?i -ingredient) (not (arm-empty))) \\
) \\
(:action putdown \\
  :parameters  (?i - ingredient) \\
  :precondition  (and (holding ?i -ingredient)) \\
  :effect (and (not (holding ?i -ingredient)) (arm-empty)) \\
) \\
(:action add \\
  :parameters  (?i - ingredient ?p - pot) \\
  :precondition (and (holding ?i -ingredient)) \\
  :effect (and (not (pot-empty ?p)) (contain ?p ?i)) \\
) \\
}
Problem file: \\
\texttt{
(define (problem threepots) \\
(:domain cooking) \\
(:objects pot1 pot2 pot3 - pot \\
ingredient1 ingredient2 ingredient3 ingredient4 ingredient5 ingredient6 - ingredient) \\
(:init  \\
(arm-empty) \\
(pot-empty pot1) \\
(pot-empty pot2) \\
(pot-empty pot3) \\
) \\
(:goal \\
(and  \\
(contain pot1 ingredient1) \\
(contain pot1 ingredient4) \\
(contain pot2 ingredient2) \\
(contain pot2 ingredient3) \\
(contain pot2 ingredient5) \\
(contain pot2 ingredient6) \\
(contain pot3 ingredient1) \\
(contain pot3 ingredient4) \\
(contain pot3 ingredient5)) \\
) 
}
\end{tcolorbox}

\begin{tcolorbox}[title={Example 2},breakable]
\textbf{Question:}
I have 4 pots and 6 different ingredients. Each ingredient can only be picked up once. Your goal is to add ingredients to pots by following the receipts: pot1 contains ingredient2, ingredient3, ingredient6. pot2 contains ingredient1, ingredient2, ingredient5. pot3 contains ingredient5, ingredient6. pot4 contains ingredient2, ingredient4, ingredient6.

\textbf{Answer:}

Domain file: \\
\texttt{
(define (domain cooking) \\
(:types pot ingredient) \\
(:predicates \\
(arm-empty) \\
(pot-empty ?p - pot) \\
(holding ?i - ingredient) \\
(contain ?p - pot ?i - ingredient) \\
) \\
(:action pick \\
  :parameters (?i - ingredient) \\
  :precondition (and (arm-empty)) \\
  :effect (and (holding ?i -ingredient) (not (arm-empty))) \\
) \\
(:action putdown \\
  :parameters  (?i - ingredient) \\
  :precondition  (and (holding ?i -ingredient)) \\
  :effect (and (not (holding ?i -ingredient)) (arm-empty)) \\
) \\
(:action add \\
  :parameters  (?i - ingredient ?p - pot) \\
  :precondition (and (holding ?i -ingredient)) \\
  :effect (and (not (pot-empty ?p)) (contain ?p ?i)) \\
) \\
}
Problem file: \\
\texttt{
(define (problem fourpots) \\
(:domain cooking) \\
(:objects pot1 pot2 pot3 pot4 - pot \\
ingredient1 ingredient2 ingredient3 ingredient4 ingredient5 ingredient6 - ingredient) \\
(:init  \\
(arm-empty) \\
(pot-empty pot1) \\
(pot-empty pot2) \\
(pot-empty pot3) \\
(pot-empty pot4) \\
) \\
(:goal \\
(and  \\
(contain pot1 ingredient2) \\
(contain pot1 ingredient3) \\
(contain pot1 ingredient6) \\
(contain pot2 ingredient1) \\
(contain pot2 ingredient2) \\
(contain pot2 ingredient5) \\
(contain pot3 ingredient5) \\
(contain pot3 ingredient6) \\
(contain pot4 ingredient2) \\
(contain pot4 ingredient4) \\
(contain pot4 ingredient6)) \\
) 
}
\end{tcolorbox}

\begin{tcolorbox}[title={Example 3},breakable]
\textbf{Question:}
I have 5 pots and 6 different ingredients. Each ingredient can only be picked up once. Your goal is to add ingredients to pots by following the receipts: pot1 contains ingredient3, ingredient4, ingredient5. pot2 contains ingredient1, ingredient2, ingredient3, ingredient6. pot3 contains ingredient2, ingredient3, ingredient5. pot4 contains ingredient1, ingredient6. pot5 contains ingredient1, ingredient3, ingredient4, ingredient5.

\textbf{Answer:}

Domain file: \\
\texttt{
(define (domain cooking) \\
(:types pot ingredient) \\
(:predicates  \\
(arm-empty) \\
(pot-empty ?p - pot) \\
(holding ?i - ingredient) \\
(contain ?p - pot ?i - ingredient) \\
) \\
(:action pick \\
  :parameters (?i - ingredient) \\
  :precondition (and (arm-empty)) \\
  :effect (and (holding ?i -ingredient) (not (arm-empty))) \\
) \\
(:action putdown \\
  :parameters  (?i - ingredient) \\
  :precondition  (and (holding ?i -ingredient)) \\
  :effect (and (not (holding ?i -ingredient)) (arm-empty)) \\
) \\
(:action add \\
  :parameters  (?i - ingredient ?p - pot) \\
  :precondition (and (holding ?i -ingredient)) \\
  :effect (and (not (pot-empty ?p)) (contain ?p ?i)) \\
) \\
}
Problem file: \\
\texttt{
(define (problem fivepots) \\
(:domain cooking) \\
(:objects pot1 pot2 pot3 pot4 pot5 - pot \\
ingredient1 ingredient2 ingredient3 ingredient4 ingredient5 ingredient6 - ingredient) \\
(:init  \\
(arm-empty) \\
(pot-empty pot1) \\
(pot-empty pot2) \\
(pot-empty pot3) \\
(pot-empty pot4) \\
(pot-empty pot5) \\
) \\
(:goal \\
(and  \\
(contain pot1 ingredient3) \\
(contain pot1 ingredient4) \\
(contain pot1 ingredient5) \\
(contain pot2 ingredient2) \\
(contain pot2 ingredient3) \\
(contain pot2 ingredient6) \\
(contain pot3 ingredient2) \\
(contain pot3 ingredient3) \\
(contain pot3 ingredient5) \\
(contain pot4 ingredient1) \\
(contain pot4 ingredient6) \\
(contain pot5 ingredient1) \\
(contain pot5 ingredient3) \\
(contain pot5 ingredient4) \\
(contain pot5 ingredient5)) \\
)
}
\end{tcolorbox}

\paragraph{Ball Moving:} 
For the Ball Moving planning domain, we use a total of three few-shot examples, which are given as follows.

\begin{tcolorbox}[title={Example 1},breakable]
\textbf{Question:}
I have 3 balls within 4 rooms. Initially: Robot is in room2. Ball ball1 is in room3. Ball ball2 is in room2. Ball ball3 is in room4. Your goal is to move the balls to specific rooms: ball1 in room1, ball2 in room2, and ball3 in room3.

\textbf{Answer:}

Domain file: \\
\texttt{
(define (domain ballmoving) \\ 
(:types room ball robot) \\
(:predicates  \\
(robot-at ?r - robot ?x - room) \\
(at ?b - ball ?x - room) \\
(arm-empty) \\
(carry ?r -robot ?b - ball) \\
) \\
(:action move \\
  :parameters (?r - robot ?from ?to - room) \\
  :precondition (and (robot-at ?r ?from)) \\
  :effect (and (robot-at ?r ?to) (not (robot-at ?r ?from))) \\
) \\
(:action pick \\
  :parameters  (?r - robot ?b - ball ?room - room) \\
  :precondition  (and (at ?b ?room) (robot-at ?r ?room) (arm-empty)) \\
  :effect (and (carry ?r ?b) (not (arm-empty)) (not (at ?b ?room)) \\
) \\
(:action drop \\
  :parameters  (?r - robot ?b -ball ?room -room) \\ 
  :precondition (and (carry ?r ?b) (robot-at ?r ?room)) \\
  :effect (and (arm-empty) (at ?b ?room) (not (carry ?r ?b))) \\
) \\
}
Problem file: \\
\texttt{
(define (problem threeballs) \\
(:domain ballmoving) \\
(:objects robot1 - robot \\
room1 room2 room3 room4 - room \\
ball1 ball2 ball3 - ball) \\
(:init  \\
(arm-empty) \\
(robot-at robot1 room2) \\
(at ball1 room3) \\
(at ball2 room2) \\
(at ball3 room4)  \\
) \\
(:goal \\
(and  \\
(at ball1 room1) \\
(at ball2 room2) \\
(at ball3 room3)) \\
)
}
\end{tcolorbox}

\begin{tcolorbox}[title={Example 2},breakable]
\textbf{Question:}
I have 4 balls within 4 rooms. Initially: Robot is in room3. Ball ball1 is in room1. Ball ball2 is in room3. Ball ball3 is in room1. Ball ball4 is in room2. Your goal is to move the balls to specific rooms: ball1 in room3, ball2 in room2, ball3 in room4, and ball4 in room4.

\textbf{Answer:}

Domain file: \\
\texttt{
(define (domain ballmoving) \\
(:types room ball robot) \\
(:predicates  \\
(robot-at ?r - robot ?x - room) \\
(at ?b - ball ?x - room) \\
(arm-empty) \\
(carry ?r -robot ?b - ball) \\
) \\
(:action move \\
  :parameters (?r - robot ?from ?to - room)\\
  :precondition (and (robot-at ?r ?from))\\
  :effect (and (robot-at ?r ?to) (not (robot-at ?r ?from)))\\
)\\
(:action pick\\
  :parameters  (?r - robot ?b - ball ?room - room)\\
  :precondition  (and (at ?b ?room) (robot-at ?r ?room) (arm-empty))\\
  :effect (and (carry ?r ?b) (not (arm-empty)) (not (at ?b ?room))\\
)\\
(:action drop\\
  :parameters  (?r - robot ?b -ball ?room -room)\\
  :precondition (and (carry ?r ?b) (robot-at ?r ?room))\\
  :effect (and (arm-empty) (at ?b ?room) (not (carry ?r ?b)))\\
)\\
}
Problem file:\\
\texttt{
(define (problem fourballs)\\
(:domain ballmoving)\\
(:objects robot1 - robot\\
room1 room2 room3 room4 - room\\
ball1 ball2 ball3 ball4 - ball)\\
(:init \\
(arm-empty)\\
(robot-at robot1 room3)\\
(at ball1 room1)\\
(at ball2 room3)\\
(at ball3 room1)\\
(at ball4 room2)\\
)\\
(:goal\\
(and \\
(at ball1 room3)\\
(at ball2 room2)\\
(at ball3 room4)\\
(at ball4 room4))\\
)
}
\end{tcolorbox}

\begin{tcolorbox}[title={Example 3},breakable]
\textbf{Question:}
I have 5 balls within 4 rooms. Initially: Robot is in room2. Ball ball1 is in room1. Ball ball2 is in room2. Ball ball3 is in room4. Ball ball4 is in room3. Ball ball5 is in room4. Your goal is to move the balls to specific rooms: ball1 in room1, ball2 in room1, ball3 in room4, ball4 in room2, and ball5 in room1.

\textbf{Answer:}

Domain file:\\
\texttt{
(define (domain ballmoving)\\
(:types room ball robot)\\
(:predicates \\
(robot-at ?r - robot ?x - room)\\
(at ?b - ball ?x - room)\\
(arm-empty)\\
(carry ?r -robot ?b - ball)\\
)\\
(:action move\\
  :parameters (?r - robot ?from ?to - room)\\
  :precondition (and (robot-at ?r ?from))\\
  :effect (and (robot-at ?r ?to) (not (robot-at ?r ?from)))\\
)\\
(:action pick\\
  :parameters  (?r - robot ?b - ball ?room - room)\\
  :precondition  (and (at ?b ?room) (robot-at ?r ?room) (arm-empty))\\
  :effect (and (carry ?r ?b) (not (arm-empty)) (not (at ?b ?room))\\
)\\
(:action drop\\
  :parameters  (?r - robot ?b -ball ?room -room)\\
  :precondition (and (carry ?r ?b) (robot-at ?r ?room))\\
  :effect (and (arm-empty) (at ?b ?room) (not (carry ?r ?b)))\\
)\\
}
Problem file:\\
\texttt{
(define (problem fiveballs)\\
(:domain ballmoving)\\
(:objects robot1 - robot\\
room1 room2 room3 room4 - room\\
ball1 ball2 ball3 ball4 ball5 - ball)\\
(:init \\
(arm-empty)\\
(robot-at robot1 room2)\\
(at ball1 room1)\\
(at ball2 room2)\\
(at ball3 room4)\\
(at ball4 room3)\\
(at ball5 room4)\\
)\\
(:goal\\
(and \\
(at ball1 room1)\\
(at ball2 room1)\\
(at ball3 room4)\\
(at ball4 room2)\\
(at ball5 room1))\\
)
}
\end{tcolorbox}

\paragraph{Blocksworld:} 
For the Blocksworld planning domain, we use a total of three few-shot examples, which are given as follows.

\begin{tcolorbox}[title={Example 1},breakable]
\textbf{Question:}
I have 3 blocks. Initially: Block b1 is on the table. Block b2 is on the table. Block b3 is on top of b1. Your goal is to move the blocks such that they are stacked in the order: b1 on b2, b2 on b3, and b3 on table.

\textbf{Answer:}

Domain file:\\
\texttt{
(define (domain blocksworld)\\
(:predicates \\
(clear ?x)\\
(on ?x ?y)\\
(on-table ?x)\\
(arm-empty)\\
(holding ?x)\\
)\\
(:action pickup\\
  :parameters (?ob)\\
  :precondition (and (clear ?ob) (on-table ?ob) (arm-empty))\\
  :effect (and (holding ?ob) (not (clear ?ob)) (not (on-table ?ob)) (not (arm-empty)))\\
)\\
(:action putdown\\
  :parameters  (?ob)\\
  :precondition (holding ?ob)\\
  :effect (and (clear ?ob) (arm-empty) (on-table ?ob) (not (holding ?ob)))\\
)\\
(:action stack\\
  :parameters  (?ob ?underob)\\
  :precondition (and (clear ?underob) (holding ?ob))\\
  :effect (and (arm-empty) (clear ?ob) (on ?ob ?underob) (not (clear ?underob)) (not (holding ?ob)))\\
)\\
(:action unstack\\
  :parameters  (?ob ?underob)\\
  :precondition (and (on ?ob ?underob) (clear ?ob) (arm-empty))\\
  :effect (and (holding ?ob) (clear ?underob) (not (on ?ob ?underob)) (not (clear ?ob)) (not (arm-empty))))\\
)\\
}
Problem file:\\
\texttt{
(define (problem threeblocks)\\
(:domain blocksworld)\\
(:objects b1 b2 b3)\\
(:init \\
(arm-empty)\\
(on-table b1)\\
(on-table b2)\\
(on b3 b1)\\
(clear b2)\\
(clear b3)\\
)\\
(:goal\\
(and \\
(on b1 b2)\\
(on b2 b3)\\
(on-table b3))\\
)
}
\end{tcolorbox}

\begin{tcolorbox}[title={Example 2},breakable]
\textbf{Question:}
I have 4 blocks. Initially: Block b1 is on the table. Block b2 is on top of b4. Block b3 is on top of b1. Block b4 is on the table. Your goal is to move the blocks such that they are stacked in the order: b3 on b2, b2 on b1, b1 on b4, and b4 on table. 

\textbf{Answer:}

Domain file:\\
\texttt{
(define (domain blocksworld)\\
(:predicates \\
(clear ?x)\\
(on ?x ?y)\\
(on-table ?x)\\
(arm-empty)\\
(holding ?x)\\
)\\
(:action pickup\\
  :parameters (?ob)\\
  :precondition (and (clear ?ob) (on-table ?ob) (arm-empty))\\
  :effect (and (holding ?ob) (not (clear ?ob)) (not (on-table ?ob)) (not (arm-empty)))\\
)\\
(:action putdown\\
  :parameters  (?ob)\\
  :precondition (holding ?ob)\\
  :effect (and (clear ?ob) (arm-empty) (on-table ?ob) (not (holding ?ob)))\\
)\\
(:action stack\\
  :parameters  (?ob ?underob)\\
  :precondition (and (clear ?underob) (holding ?ob))\\
  :effect (and (arm-empty) (clear ?ob) (on ?ob ?underob) (not (clear ?underob)) (not (holding ?ob)))\\
)\\
(:action unstack\\
  :parameters  (?ob ?underob)\\
  :precondition (and (on ?ob ?underob) (clear ?ob) (arm-empty))\\
  :effect (and (holding ?ob) (clear ?underob) (not (on ?ob ?underob)) (not (clear ?ob)) (not (arm-empty))))\\
)\\
}
Problem file:\\
\texttt{
(define (problem fourblocks)\\
(:domain blocksworld)\\
(:objects b1 b2 b3 b4)\\
(:init \\
(arm-empty)\\
(on-table b1)\\
(on b2 b4)\\
(on b3 b1)\\
(on-table b4)\\
(clear b2)\\
(clear b3)\\
)\\
(:goal\\
(and\\
(on b3 b2)\\
(on b2 b1)\\
(on b1 b4)\\
(on-table b4))\\
)
}
\end{tcolorbox}

\begin{tcolorbox}[title={Example 3},breakable]
\textbf{Question:}
I have 5 blocks. Initially: Block b1 is on the table. Block b2 is on the table. Block b3 is on top of b2. Block b4 is on the table. Block b5 is on top of b4. Your goal is to move the blocks such that they are stacked in the order: b3 on b1, b1 on b4, b4 on b2, b2 on b5, and b5 on table. 

\textbf{Answer:}

Domain file:\\
\texttt{
(define (domain blocksworld)\\
(:predicates \\
(clear ?x)\\
(on ?x ?y)\\
(on-table ?x)\\
(arm-empty)\\
(holding ?x)\\
)\\
(:action pickup\\
  :parameters (?ob)\\
  :precondition (and (clear ?ob) (on-table ?ob) (arm-empty))\\
  :effect (and (holding ?ob) (not (clear ?ob)) (not (on-table ?ob)) (not (arm-empty)))\\
)\\
(:action putdown\\
  :parameters  (?ob)\\
  :precondition (holding ?ob)\\
  :effect (and (clear ?ob) (arm-empty) (on-table ?ob) (not (holding ?ob)))\\
)\\
(:action stack\\
  :parameters  (?ob ?underob)\\
  :precondition (and (clear ?underob) (holding ?ob))\\
  :effect (and (arm-empty) (clear ?ob) (on ?ob ?underob) (not (clear ?underob)) (not (holding ?ob)))\\
)\\
(:action unstack\\
  :parameters  (?ob ?underob)\\
  :precondition (and (on ?ob ?underob) (clear ?ob) (arm-empty))\\
  :effect (and (holding ?ob) (clear ?underob) (not (on ?ob ?underob)) (not (clear ?ob)) (not (arm-empty))))\\
)\\
}
Problem file:\\
\texttt{
(define (problem fiveblocks)\\
(:domain blocksworld)\\
(:objects b1 b2 b3 b4 b5)\\
(:init \\
(arm-empty)\\
(on-table b1)\\
(on-table b2)\\
(on b3 b2)\\
(on-table b4)\\
(on b5 b4)\\
(clear b1)\\
(clear b3)\\
(clear b5)\\
)\\
(:goal\\
(and\\
(on b3 b1)\\
(on b1 b4)\\
(on b4 b2)\\
(on b2 b5)\\
(on-table b5))\\
)
}
\end{tcolorbox}

\subsection{Planner}
\label{sec.planner_example}

\paragraph{Cooking:} 
For the Cooking planning domain, we use a total of three few-shot examples, which are given as follows.

\begin{tcolorbox}[title={Example 1},breakable]
\textbf{Question:}

Domain file:\\
\texttt{
(define (domain cooking)\\
(:types pot ingredient)\\
(:predicates \\
(arm-empty)\\
(pot-empty ?p - pot)\\
(holding ?i - ingredient)\\
(contain ?p - pot ?i - ingredient)\\
)\\
(:action pick\\
  :parameters (?i - ingredient)\\
  :precondition (and (arm-empty))\\
  :effect (and (holding ?i -ingredient) (not (arm-empty)))\\
)\\
(:action putdown\\
  :parameters  (?i - ingredient)\\
  :precondition  (and (holding ?i -ingredient))\\
  :effect (and (not (holding ?i -ingredient)) (arm-empty))\\
)\\
(:action add\\
  :parameters  (?i - ingredient ?p - pot)\\
  :precondition (and (holding ?i -ingredient))\\
  :effect (and (not (pot-empty ?p)) (contain ?p ?i))\\
)\\
}
Problem file:\\
\texttt{
(define (problem threepots)\\
(:domain cooking)\\
(:objects pot1 pot2 pot3 - pot\\
ingredient1 ingredient2 ingredient3 ingredient4 ingredient5 ingredient6 - ingredient)\\
(:init \\
(arm-empty)\\
(pot-empty pot1)\\
(pot-empty pot2)\\
(pot-empty pot3)\\
)\\
(:goal\\
(and \\
(contain pot1 ingredient1)\\
(contain pot1 ingredient4)\\
(contain pot2 ingredient2)\\
(contain pot2 ingredient3)\\
(contain pot2 ingredient5)\\
(contain pot2 ingredient6)\\
(contain pot3 ingredient1)\\
(contain pot3 ingredient4)\\
(contain pot3 ingredient5))\\
)
}

\textbf{Answer:}

Action Sequence: \\
Step 1: ingredient1 is needed in pot1, pot3 \\
\texttt{(pick ingredient1)\\
(add ingredient1 pot1)\\
(add ingredient1 pot3)\\
(putdown ingredient1)\\}
Step 2: ingredient2 is needed in pot2\\
\texttt{(pick ingredient2)\\
(add ingredient2 pot2)\\
(putdown ingredient2)\\}
Step 3: ingredient3 is needed in pot2\\
\texttt{(pick ingredient3)\\
(add ingredient3 pot2)\\
(putdown ingredient3)\\}
Step 4: ingredient4 is needed in pot1, pot3\\
\texttt{(pick ingredient4)\\
(add ingredient4 pot1)\\
(add ingredient4 pot3)\\
(putdown ingredient4)\\}
Step 5: ingredient5 is needed in pot2, pot3\\
\texttt{(pick ingredient5)\\
(add ingredient5 pot2)\\
(add ingredient5 pot3)\\
(putdown ingredient5)\\}
Step 6: ingredient6 is needed in pot2\\
\texttt{(pick ingredient6)\\
(add ingredient6 pot2)\\
(putdown ingredient6)\\}
All ingredients have been added to the corresponding pots.
\end{tcolorbox}

\begin{tcolorbox}[title={Example 2},breakable]
\textbf{Question:}

Domain file:\\
\texttt{
(define (domain cooking)\\
(:types pot ingredient)\\
(:predicates \\
(arm-empty)\\
(pot-empty ?p - pot)\\
(holding ?i - ingredient)\\
(contain ?p - pot ?i - ingredient)\\
)\\
(:action pick\\
  :parameters (?i - ingredient)\\
  :precondition (and (arm-empty))\\
  :effect (and (holding ?i -ingredient) (not (arm-empty)))\\
)\\
(:action putdown\\
  :parameters  (?i - ingredient)\\
  :precondition  (and (holding ?i -ingredient))\\
  :effect (and (not (holding ?i -ingredient)) (arm-empty))\\
)\\
(:action add\\
  :parameters  (?i - ingredient ?p - pot)\\
  :precondition (and (holding ?i -ingredient))\\
  :effect (and (not (pot-empty ?p)) (contain ?p ?i))\\
)\\
}
Problem file:\\
\texttt{
(define (problem fourpots)\\
(:domain cooking)\\
(:objects pot1 pot2 po3 pot4 - pot\\
ingredient1 ingredient2 ingredient3 ingredient4 ingredient5 ingredient6 - ingredient)\\
(:init \\
(arm-empty)\\
(pot-empty pot1)\\
(pot-empty pot2)\\
(pot-empty pot3)\\
(pot-empty pot4)\\
)\\
(:goal\\
(and \\
(contain pot1 ingredient2)\\
(contain pot1 ingredient6)\\
(contain pot2 ingredient1)\\
(contain pot2 ingredient2)\\
(contain pot2 ingredient5)\\
(contain pot3 ingredient5)\\
(contain pot3 ingredient6)\\
(contain pot4 ingredient2)\\
(contain pot4 ingredient4)\\
(contain pot4 ingredient6))\\
)\\
}

\textbf{Answer:}

Action Sequence: \\
Step 1: ingredient1 is needed in pot2\\
\texttt{
(pick ingredient1)\\
(add ingredient1 pot2)\\
(putdown ingredient1)\\}
Step 2: ingredient2 is needed in pot1, pot2, pot4\\
\texttt{
(pick ingredient2)\\
(add ingredient2 pot1)\\
(add ingredient2 pot2)\\
(add ingredient2 pot4)\\
(putdown ingredient2)\\}
Step 3: ingredient3 is not needed in any pot\\
no action needed\\
Step 4: ingredient4 is needed in pot4\\
\texttt{
(pick ingredient4)\\
(add ingredient4 pot4)\\
(putdown ingredient4)\\}
Step 5: ingredient5 is needed in pot2, pot3\\
\texttt{
(pick ingredient5)\\
(add ingredient5 pot2)\\
(add ingredient5 pot3)\\
(putdown ingredient5)\\}
Step 6: ingredient6 is needed in pot1, pot3, pot4\\
\texttt{
(pick ingredient6)\\
(add ingredient6 pot1)\\
(add ingredient6 pot3)\\
(add ingredient6 pot4)\\
(putdown ingredient6)\\}
All ingredients have been added to the corresponding pots.
\end{tcolorbox}

\begin{tcolorbox}[title={Example 3},breakable]
\textbf{Question:}

Domain file:\\
\texttt{
(define (domain cooking)\\
(:types pot ingredient)\\
(:predicates \\
(arm-empty)\\
(pot-empty ?p - pot)\\
(holding ?i - ingredient)\\
(contain ?p - pot ?i - ingredient)\\
)\\
(:action pick\\
  :parameters (?i - ingredient)\\
  :precondition (and (arm-empty))\\
  :effect (and (holding ?i -ingredient) (not (arm-empty)))\\
)\\
(:action putdown\\
  :parameters  (?i - ingredient)\\
  :precondition  (and (holding ?i -ingredient))\\
  :effect (and (not (holding ?i -ingredient)) (arm-empty))\\
)\\
(:action add\\
  :parameters  (?i - ingredient ?p - pot)\\
  :precondition (and (holding ?i -ingredient))\\
  :effect (and (not (pot-empty ?p)) (contain ?p ?i))\\
)\\
}
Problem file:\\
\texttt{
(define (problem fivepots)\\
(:domain cooking)\\
(:objects pot1 pot2 pot3 pot4 pot5 - pot\\
ingredient1 ingredient2 ingredient3 ingredient4 ingredient5 ingredient6 - ingredient)\\
(:init \\
(arm-empty)\\
(pot-empty pot1)\\
(pot-empty pot2)\\
(pot-empty pot3)\\
(pot-empty pot4)\\
(pot-empty pot5)\\
)\\
(:goal\\
(and \\
(contain pot1 ingredient3)\\
(contain pot1 ingredient4)\\
(contain pot1 ingredient5)\\
(contain pot2 ingredient1)\\
(contain pot2 ingredient2)\\
(contain pot2 ingredient3)\\
(contain pot2 ingredient6)\\
(contain pot3 ingredient2)\\
(contain pot3 ingredient3)\\
(contain pot3 ingredient5)\\
(contain pot4 ingredient1)\\
(contain pot4 ingredient6)\\
(contain pot5 ingredient1)\\
(contain pot5 ingredient3)\\
(contain pot5 ingredient4)\\
(contain pot5 ingredient5))\\
)\\
}

\textbf{Answer:}

Action Sequence:\\
Step 1: ingredient1 is needed in pot2, pot4, pot5\\
\texttt{
(pick ingredient1)\\
(add ingredient1 pot2)\\
(add ingredient1 pot4)\\
(add ingredient1 pot5)\\
(putdown ingredient1)\\}
Step 2: ingredient2 is needed in pot2, pot3\\
\texttt{
(pick ingredient2)\\
(add ingredient2 pot2)\\
(add ingredient2 pot3)\\
(putdown ingredient2)\\}
Step 3: ingredient3 is needed in pot1, pot2, pot3, pot5\\
\texttt{
(pick ingredient3)\\
(add ingredient3 pot1)\\
(add ingredient3 pot2)\\
(add ingredient3 pot3)\\
(add ingredient3 pot5)\\
(putdown ingredient3)\\}
Step 4: ingredient4 is needed in pot1, pot5\\
\texttt{
(pick ingredient4)\\
(add ingredient4 pot1)\\
(add ingredient4 pot5)\\
(putdown ingredient4)\\}
Step 5: ingredient5 is needed in pot1, pot3, pot5\\
\texttt{
(pick ingredient5)\\
(add ingredient5 pot1)\\
(add ingredient5 pot3)\\
(add ingredient5 pot5)\\
(putdown ingredient5)\\}
Step 6: ingredient6 is needed in pot2, pot4\\
\texttt{
(pick ingredient6)\\
(add ingredient6 pot2)\\
(add ingredient6 pot4)\\
(putdown ingredient6)\\}
All ingredients have been added to the corresponding pots.
\end{tcolorbox}

\paragraph{Ball Moving:} 
For the Ball Moving planning domain, we use a total of three few-shot examples, which are given as follows.

\begin{tcolorbox}[title={Example 1},breakable]
\textbf{Question:}

Domain file:\\
\texttt{
(define (domain ballmoving)\\
(:types room ball robot)\\
(:predicates \\
(robot-at ?r - robot ?x - room)\\
(at ?b - ball ?x - room)\\
(arm-empty)\\
(carry ?r -robot ?b - ball)\\
)\\
(:action move\\
  :parameters (?r - robot ?from ?to - room)\\
  :precondition (and (robot-at ?r ?from))\\
  :effect (and (robot-at ?r ?to) (not (robot-at ?r ?from)))\\
)\\
(:action pick\\
  :parameters  (?r - robot ?b - ball ?room - room)\\
  :precondition  (and (at ?b ?room) (robot-at ?r ?room) (arm-empty))\\
  :effect (and (carry ?r ?b) (not (arm-empty)) (not (at ?b ?room))\\
)\\
(:action drop\\
  :parameters  (?r - robot ?b -ball ?room -room)\\
  :precondition (and (carry ?r ?b) (robot-at ?r ?room))\\
  :effect (and (arm-empty) (at ?b ?room) (not (carry ?r ?b)))\\
)\\
}
Problem file:\\
\texttt{
(define (problem threeballs)\\
(:domain ballmoving)\\
(:objects robot1 - robot\\
room1 room2 room3 room4 - room\\
ball1 ball2 ball3 - ball)\\
(:init \\
(arm-empty)\\
(robot-at robot1 room2)\\
(at ball1 room3)\\
(at ball2 room2)\\
(at ball3 room4)\\
)\\
(:goal\\
(and \\
(at ball1 room1)\\
(at ball2 room2)\\
(at ball3 room3))\\
)\\
}

\textbf{Answer:}

Action Sequence:\\
Step 1: \\
robot1 is now in room2. \\
\texttt{(at ball1 room3)} with goal \texttt{(at ball1 room1)}\\
\texttt{(at ball2 room2)} with goal \texttt{(at ball2 room2)}\\
\texttt{(at ball3 room4)} with goal \texttt{(at ball3 room3)}\\
ball1 is unsatisfied, ball2 is satisfied, ball3 is unsatisfied.\\
All balls that are in the same room with robot1 are satisfied.\\
Go to complete next unsatisfied ball (from first to last): ball1 \\
\texttt{(move robot1 room2 room3)\\
(pick ball1 room3)\\
(move robot1 room3 room1)\\
(drop ball1 room1)\\}
Step 2:\\
robot1 is now in room1. \\
\texttt{(at ball1 room1)} with goal \texttt{(at ball1 room1)}\\
\texttt{(at ball2 room2)} with goal \texttt{(at ball2 room2)}\\
\texttt{(at ball3 room4)} with goal \texttt{(at ball3 room3)}\\
ball1 is satisfied, ball2 is satisfied, ball3 is unsatisfied.\\
All balls that are in the same room with robot1 are satisfied.\\
Go to complete next unsatisfied ball (from first to last): ball3 \\
\texttt{
(move robot1 room1 room4)\\
(pick ball3 room4)\\
(move robot1 room4 room3)\\
(drop ball3 room3)\\}
Step 3:\\
robot1 is now in room3.\\
\texttt{(at ball1 room1)} with goal \texttt{(at ball1 room1)}\\
\texttt{(at ball2 room2)} with goal \texttt{(at ball2 room2)}\\
\texttt{(at ball3 room3)} with goal \texttt{(at ball3 room3)}\\
ball1 is satisfied, ball2 is satisfied, ball3 is satisfied.\\
All balls are satisfied.
\end{tcolorbox}

\begin{tcolorbox}[title={Example 2},breakable]
\textbf{Question:}

Domain file:\\
\texttt{
(define (domain ballmoving)\\
(:types room ball robot)\\
(:predicates \\
(robot-at ?r - robot ?x - room)\\
(at ?b - ball ?x - room)\\
(arm-empty)\\
(carry ?r -robot ?b - ball)\\
)\\
(:action move\\
  :parameters (?r - robot ?from ?to - room)\\
  :precondition (and (robot-at ?r ?from))\\
  :effect (and (robot-at ?r ?to) (not (robot-at ?r ?from)))\\
)\\
(:action pick\\
  :parameters  (?r - robot ?b - ball ?room - room)\\
  :precondition  (and (at ?b ?room) (robot-at ?r ?room) (arm-empty))\\
  :effect (and (carry ?r ?b) (not (arm-empty)) (not (at ?b ?room))\\
)\\
(:action drop\\
  :parameters  (?r - robot ?b -ball ?room -room)\\
  :precondition (and (carry ?r ?b) (robot-at ?r ?room))\\
  :effect (and (arm-empty) (at ?b ?room) (not (carry ?r ?b)))\\
)\\
}
Problem file:\\
\texttt{
(define (problem fourballs)\\
(:domain ballmoving)\\
(:objects robot1 - robot\\
room1 room2 room3 room4 - room\\
ball1 ball2 ball3 ball4 - ball)\\
(:init \\
(arm-empty)\\
(robot-at robot1 room3)\\
(at ball1 room1)\\
(at ball2 room3)\\
(at ball3 room1)\\
(at ball4 room2)\\
)\\
(:goal\\
(and \\
(at ball1 room3)\\
(at ball2 room2)\\
(at ball3 room4)\\
(at ball4 room4))\\
)\\
}

\textbf{Answer:}

Action Sequence:\\
Step 1: \\
robot1 is now in room3. \\
\texttt{(at ball1 room1)} with goal \texttt{(at ball1 room3)}\\
\texttt{(at ball2 room3)} with goal \texttt{(at ball2 room2)}\\
\texttt{(at ball3 room1)} with goal \texttt{(at ball3 room4)}\\
\texttt{(at ball4 room2)} with goal \texttt{(at ball4 room4)}\\
ball1 is unsatisfied, ball2 is unsatisfied, ball3 is unsatisfied, ball4 is unsatisfied.\\
ball2 is in the same room with the robot1 and is unsatisfied.\\
Go to complete ball2 \\
\texttt{
(pick ball2 room3)\\
(move robot1 room3 room2)\\
(drop ball2 room2)\\}
Step 2:\\
robot1 is now in room2. \\
\texttt{(at ball1 room1)} with goal \texttt{(at ball1 room3)}\\
\texttt{(at ball2 room2)} with goal \texttt{(at ball2 room2)}\\
\texttt{(at ball3 room1)} with goal \texttt{(at ball3 room4)}\\
\texttt{(at ball4 room2)} with goal \texttt{(at ball4 room4)}\\
ball1 is unsatisfied, ball2 is satisfied, ball3 is unsatisfied, ball4 is unsatisfied.\\
ball4 is in the same room with the robot1 and is unsatisfied.\\
Go to complete ball4 \\
\texttt{
(pick ball4 room2)\\
(move robot1 room2 room4)\\
(drop ball4 room4)\\}
Step 3:\\
robot1 is now in room4.\\
\texttt{(at ball1 room1)} with goal \texttt{(at ball1 room3)}\\
\texttt{(at ball2 room2)} with goal \texttt{(at ball2 room2)}\\
\texttt{(at ball3 room1)} with goal \texttt{(at ball3 room4)}\\
\texttt{(at ball4 room4)} with goal \texttt{(at ball4 room4)}\\
ball1 is unsatisfied, ball2 is satisfied, ball3 is unsatisfied, ball4 is satisfied.\\
All balls that are in the same room with robot1 are satisfied.\\
Go to complete next unsatisfied ball (from first to last): ball1 \\
\texttt{
(move robot1 room4 room1)\\
(pick ball1 room1)\\
(move robot1 room1 room3)\\
(drop ball1 room3)\\}
Step 4:\\
robot1 is now in room3.\\
\texttt{(at ball1 room3)} with goal \texttt{(at ball1 room3)}\\
\texttt{(at ball2 room2)} with goal \texttt{(at ball2 room2)}\\
\texttt{(at ball3 room1)} with goal \texttt{(at ball3 room4)}\\
\texttt{(at ball4 room4)} with goal \texttt{(at ball4 room4)}\\
ball1 is satisfied, ball2 is satisfied, ball3 is unsatisfied, ball4 is satisfied.\\
All balls that are in the same room with robot1 are satisfied.\\
Go to complete next unsatisfied ball (from first to last): ball3 \\
\texttt{
(move robot1 room3 room1)\\
(pick ball3 room1)\\
(move robot1 room1 room4)\\
(drop ball3 room4)\\}
Step 5:\\
robot1 is now in room4.\\
\texttt{(at ball1 room3)} with goal \texttt{(at ball1 room3)}\\
\texttt{(at ball2 room2)} with goal \texttt{(at ball2 room2)}\\
\texttt{(at ball3 room4)} with goal \texttt{(at ball3 room4)}\\
\texttt{(at ball4 room4)} with goal \texttt{(at ball4 room4)}\\
ball1 is satisfied, ball2 is satisfied, ball3 is satisfied, ball4 is satisfied.\\
All balls are satisfied.
\end{tcolorbox}

\begin{tcolorbox}[title={Example 3},breakable]
\textbf{Question:}

Domain file:\\
\texttt{
(define (domain ballmoving)\\
(:types room ball robot)\\
(:predicates \\
(robot-at ?r - robot ?x - room)\\
(at ?b - ball ?x - room)\\
(arm-empty)\\
(carry ?r -robot ?b - ball)\\
)\\
(:action move\\
  :parameters (?r - robot ?from ?to - room)\\
  :precondition (and (robot-at ?r ?from))\\
  :effect (and (robot-at ?r ?to) (not (robot-at ?r ?from)))\\
)\\
(:action pick\\
  :parameters  (?r - robot ?b - ball ?room - room)\\
  :precondition  (and (at ?b ?room) (robot-at ?r ?room) (arm-empty))\\
  :effect (and (carry ?r ?b) (not (arm-empty)) (not (at ?b ?room))\\
)\\
(:action drop\\
  :parameters  (?r - robot ?b -ball ?room -room)\\
  :precondition (and (carry ?r ?b) (robot-at ?r ?room))\\
  :effect (and (arm-empty) (at ?b ?room) (not (carry ?r ?b)))\\
)\\
}
Problem file:\\
\texttt{
(define (problem fiveballs)\\
(:domain ballmoving)\\
(:objects robot1 - robot\\
room1 room2 room3 room4 - room\\
ball1 ball2 ball3 ball4 ball5 - ball)\\
(:init \\
(arm-empty)\\
(robot-at robot1 room2)\\
(at ball1 room1)\\
(at ball2 room2)\\
(at ball3 room4)\\
(at ball4 room3)\\
(at ball5 room4)\\
)\\
(:goal\\
(and \\
(at ball1 room1)\\
(at ball2 room1)\\
(at ball3 room4)\\
(at ball4 room2)\\
(at ball5 room1))\\
)\\
}

\textbf{Answer:}

Action Sequence:\\
Step 1: \\
robot1 is now in room2. \\
\texttt{(at ball1 room1)} with goal \texttt{(at ball1 room1)}\\
\texttt{(at ball2 room2)} with goal \texttt{(at ball2 room1)}\\
\texttt{(at ball3 room4)} with goal \texttt{(at ball3 room4)}\\
\texttt{(at ball4 room3)} with goal \texttt{(at ball4 room2)}\\
\texttt{(at ball5 room4)} with goal \texttt{(at ball5 room1)}\\
ball1 is satisfied, ball2 is unsatisfied, ball3 is satisfied, ball4 is unsatisfied, ball5 is unsatisfied.\\
ball2 is in the same room with the robot1 and is unsatisfied.\\
Go to complete ball2 \\
\texttt{
(pick ball2 room2)\\
(move robot1 room2 room1)\\
(drop ball2 room1)\\}
Step 2:\\
robot1 is now in room1. \\
\texttt{(at ball1 room1)} with goal \texttt{(at ball1 room1)}\\
\texttt{(at ball2 room1)} with goal \texttt{(at ball2 room1)}\\
\texttt{(at ball3 room4)} with goal \texttt{(at ball3 room4)}\\
\texttt{(at ball4 room3)} with goal \texttt{(at ball4 room2)}\\
\texttt{(at ball5 room4)} with goal \texttt{(at ball5 room1)}\\
ball1 is satisfied, ball2 is satisfied, ball3 is satisfied, ball4 is unsatisfied, ball5 is unsatisfied.\\
All balls that are in the same room with robot1 are satisfied.\\
Go to complete next unsatisfied ball (from first to last): ball4\\
\texttt{
(move robot1 room1 room3)\\
(pick ball4 room3)\\
(move robot1 room3 room2)\\
(drop ball4 room2)\\}
Step 3:\\
robot1 is now in room2.\\
\texttt{(at ball1 room1)} with goal \texttt{(at ball1 room1)}\\
\texttt{(at ball2 room1)} with goal \texttt{(at ball2 room1)}\\
\texttt{(at ball3 room4)} with goal \texttt{(at ball3 room4)}\\
\texttt{(at ball4 room2)} with goal \texttt{(at ball4 room2)}\\
\texttt{(at ball5 room4)} with goal \texttt{(at ball5 room1)}\\
ball1 is satisfied, ball2 is satisfied, ball3 is satisfied, ball4 is satisfied, ball5 is unsatisfied.\\
All balls that are in the same room with robot1 are satisfied.\\
Go to complete next unsatisfied ball (from first to last): ball5 \\
\texttt{
(move robot1 room2 room4)\\
(pick ball5 room4)\\
(move robot1 room4 room1)\\
(drop ball5 room1)\\}
Step 4:\\
robot1 is now in room1.\\
\texttt{(at ball1 room1)} with goal \texttt{(at ball1 room1)}\\
\texttt{(at ball2 room1)} with goal \texttt{(at ball2 room1)}\\
\texttt{(at ball3 room4)} with goal \texttt{(at ball3 room4)}\\
\texttt{(at ball4 room2)} with goal \texttt{(at ball4 room2)}\\
\texttt{(at ball5 room1)} with goal \texttt{(at ball5 room1)}\\
ball1 is satisfied, ball2 is satisfied, ball3 is satisfied, ball4 is satisfied, ball5 is satisfied.\\
All balls are satisfied.
\end{tcolorbox}

\paragraph{Blocksworld:} 
For the Blocksworld planning domain, we use a total of five few-shot examples, which are given as follows.

\begin{tcolorbox}[title={Example 1},breakable]
\textbf{Question:}

Domain file:\\
\texttt{
(define (domain blocksworld)\\
(:predicates \\
(clear ?x)\\
(on ?x ?y)\\
(on-table ?x)\\
(arm-empty)\\
(holding ?x)\\
)\\
(:action pickup\\
  :parameters (?ob)\\
  :precondition (and (clear ?ob) (on-table ?ob) (arm-empty))\\
  :effect (and (holding ?ob) (not (clear ?ob)) (not (on-table ?ob)) (not (arm-empty)))\\
)\\
(:action putdown\\
  :parameters  (?ob)\\
  :precondition (holding ?ob)\\
  :effect (and (clear ?ob) (arm-empty) (on-table ?ob) (not (holding ?ob)))\\
)\\
(:action stack\\
  :parameters  (?ob ?underob)\\
  :precondition (and (clear ?underob) (holding ?ob))\\
  :effect (and (arm-empty) (clear ?ob) (on ?ob ?underob) (not (clear ?underob)) (not (holding ?ob)))\\
)\\
(:action unstack\\
  :parameters  (?ob ?underob)\\
  :precondition (and (on ?ob ?underob) (clear ?ob) (arm-empty))\\
  :effect (and (holding ?ob) (clear ?underob) (not (on ?ob ?underob)) (not (clear ?ob)) (not (arm-empty))))\\
)\\
}
Problem file:\\
\texttt{
(define (problem threeblocks)\\
(:domain blocksworld)\\
(:objects b1 b2 b3)\\
(:init \\
(arm-empty)\\
(on-table b1)\\
(on-table b2)\\
(on b3 b1)\\
(clear b2)\\
(clear b3)\\
)\\
(:goal\\
(and \\
(on b1 b2)\\
(on b2 b3)\\
(on-table b3))\\
)\\
}

\textbf{Answer:}

Action Sequence:\\
We need to build the blocks from bottom to top.\\
Third goal: b3 on table\\
\texttt{
(unstack b3 b1)\\
(putdown b3)\\}
Second goal: b2 on b3\\
\texttt{(pickup b2)\\
(stack b2 b3)\\}
First goal: b1 on b2\\
\texttt{(pickup b1)\\
(stack b1 b2)}
\end{tcolorbox}

\begin{tcolorbox}[title={Example 2},breakable]
\textbf{Question:}

Domain file:\\
\texttt{
(define (domain blocksworld)\\
(:predicates \\
(clear ?x)\\
(on ?x ?y)\\
(on-table ?x)\\
(arm-empty)\\
(holding ?x)\\
)\\
(:action pickup\\
  :parameters (?ob)\\
  :precondition (and (clear ?ob) (on-table ?ob) (arm-empty))\\
  :effect (and (holding ?ob) (not (clear ?ob)) (not (on-table ?ob)) (not (arm-empty)))\\
)\\
(:action putdown\\
  :parameters  (?ob)\\
  :precondition (holding ?ob)\\
  :effect (and (clear ?ob) (arm-empty) (on-table ?ob) (not (holding ?ob)))\\
)\\
(:action stack\\
  :parameters  (?ob ?underob)\\
  :precondition (and (clear ?underob) (holding ?ob))\\
  :effect (and (arm-empty) (clear ?ob) (on ?ob ?underob) (not (clear ?underob)) (not (holding ?ob)))\\
)\\
(:action unstack\\
  :parameters  (?ob ?underob)\\
  :precondition (and (on ?ob ?underob) (clear ?ob) (arm-empty))\\
  :effect (and (holding ?ob) (clear ?underob) (not (on ?ob ?underob)) (not (clear ?ob)) (not (arm-empty))))\\
)\\
}
Problem file:\\
\texttt{
(define (problem fourblocks)\\
(:domain blocksworld)\\
(:objects b1 b2 b3 b4)\\
(:init \\
(arm-empty)\\
(on-table b1)\\
(on b2 b4)\\
(on b3 b1)\\
(on-table b4)\\
(clear b2)\\
(clear b3)\\
)\\
(:goal\\
(and\\
(on b3 b2)\\
(on b2 b1)\\
(on b1 b4)\\
(on-table b4))\\
)\\
}

\textbf{Answer:}

Action Sequence:\\
We need to build the blocks from bottom to top.\\
Fourth goal: b4 on table\\
Already satisfied in initial configuration\\
Third goal: b1 on b4\\
\texttt{(unstack b2 b4)\\
(putdown b2)\\
(unstack b3 b1)\\
(putdown b3)\\
(pickup b1)\\
(stack b1 b4)\\}
Second goal: b2 on b1\\
\texttt{(pickup b2)\\
(stack b2 b1)\\}
First goal: b3 on b2\\
\texttt{(pickup b3)\\
(stack b3 b2)}
\end{tcolorbox}

\begin{tcolorbox}[title={Example 3},breakable]
\textbf{Question:}

Domain file:\\
\texttt{
(define (domain blocksworld)\\
(:predicates \\
(clear ?x)\\
(on ?x ?y)\\
(on-table ?x)\\
(arm-empty)\\
(holding ?x)\\
)\\
(:action pickup\\
  :parameters (?ob)\\
  :precondition (and (clear ?ob) (on-table ?ob) (arm-empty))\\
  :effect (and (holding ?ob) (not (clear ?ob)) (not (on-table ?ob)) (not (arm-empty)))\\
)\\
(:action putdown\\
  :parameters  (?ob)\\
  :precondition (holding ?ob)\\
  :effect (and (clear ?ob) (arm-empty) (on-table ?ob) (not (holding ?ob)))\\
)\\
(:action stack\\
  :parameters  (?ob ?underob)\\
  :precondition (and (clear ?underob) (holding ?ob))\\
  :effect (and (arm-empty) (clear ?ob) (on ?ob ?underob) (not (clear ?underob)) (not (holding ?ob)))\\
)\\
(:action unstack\\
  :parameters  (?ob ?underob)\\
  :precondition (and (on ?ob ?underob) (clear ?ob) (arm-empty))\\
  :effect (and (holding ?ob) (clear ?underob) (not (on ?ob ?underob)) (not (clear ?ob)) (not (arm-empty))))\\
)\\
}
Problem file:\\
\texttt{
(define (problem fiveblocks)\\
(:domain blocksworld)\\
(:objects b1 b2 b3 b4 b5)\\
(:init \\
(arm-empty)\\
(on-table b1)\\
(on-table b2)\\
(on b3 b2)\\
(on-table b4)\\
(on b5 b4)\\
(clear b1)\\
(clear b3)\\
(clear b5)\\
)\\
(:goal\\
(and\\
(on b3 b1)\\
(on b1 b4)\\
(on b4 b2)\\
(on b2 b5)\\
(on-table b5))\\
)\\
}

\textbf{Answer:}

Action Sequence:\\
We need to build the blocks from bottom to top.\\
Fifth goal: b5 on table\\
\texttt{(unstack b5 b4)\\
(putdown b5)\\}
Fourth goal: b2 on b5\\
\texttt{(unstack b3 b2)\\
(putdown b3)\\
(pickup b2)\\
(stack b2 b5)\\}
Third goal: b4 on b2\\
\texttt{(pickup b4)\\
(stack b4 b2)\\}
Second goal: b1 on b4\\
\texttt{(pickup b1)\\
(stack b1 b4)\\}
First goal: b3 on b1\\
\texttt{(pickup b3)\\
(stack b3 b1)}
\end{tcolorbox}

\begin{tcolorbox}[title={Example 4},breakable]
\textbf{Question:}

Domain file:\\
\texttt{
(define (domain blocksworld)\\
(:predicates \\
(clear ?x)\\
(on ?x ?y)\\
(on-table ?x)\\
(arm-empty)\\
(holding ?x)\\
)\\
(:action pickup\\
  :parameters (?ob)\\
  :precondition (and (clear ?ob) (on-table ?ob) (arm-empty))\\
  :effect (and (holding ?ob) (not (clear ?ob)) (not (on-table ?ob)) (not (arm-empty)))\\
)\\
(:action putdown\\
  :parameters  (?ob)\\
  :precondition (holding ?ob)\\
  :effect (and (clear ?ob) (arm-empty) (on-table ?ob) (not (holding ?ob)))\\
)\\
(:action stack\\
  :parameters  (?ob ?underob)\\
  :precondition (and (clear ?underob) (holding ?ob))\\
  :effect (and (arm-empty) (clear ?ob) (on ?ob ?underob) (not (clear ?underob)) (not (holding ?ob)))\\
)\\
(:action unstack\\
  :parameters  (?ob ?underob)\\
  :precondition (and (on ?ob ?underob) (clear ?ob) (arm-empty))\\
  :effect (and (holding ?ob) (clear ?underob) (not (on ?ob ?underob)) (not (clear ?ob)) (not (arm-empty))))\\
)\\
}
Problem file:\\
\texttt{
(define (problem threeblocks)\\
(:domain blocksworld)\\
(:objects b1 b2 b3)\\
(:init \\
(arm-empty)\\
(on b1 b3)\\
(on-table b2)\\
(on-table b3)\\
(clear b1)\\
(clear b2)\\
)\\
(:goal\\
(and \\
(on b2 b1)\\
(on b1 b3)\\
(on-table b3))\\
)\\
}

\textbf{Answer:}

Action Sequence:\\
We need to build the blocks from bottom to top.\\
Third goal: b3 on table\\
Already satisfied in initial configuration\\
Second goal: b1 on b3\\
Already satisfied in initial configuration\\
First goal: b2 on b1\\
\texttt{(pickup b2)\\
(stack b2 b1)}
\end{tcolorbox}

\begin{tcolorbox}[title={Example 5},breakable]
\textbf{Question:}

Domain file:\\
\texttt{
(define (domain blocksworld)\\
(:predicates \\
(clear ?x)\\
(on ?x ?y)\\
(on-table ?x)\\
(arm-empty)\\
(holding ?x)\\
)\\
(:action pickup\\
  :parameters (?ob)\\
  :precondition (and (clear ?ob) (on-table ?ob) (arm-empty))\\
  :effect (and (holding ?ob) (not (clear ?ob)) (not (on-table ?ob)) (not (arm-empty)))\\
)\\
(:action putdown\\
  :parameters  (?ob)\\
  :precondition (holding ?ob)\\
  :effect (and (clear ?ob) (arm-empty) (on-table ?ob) (not (holding ?ob)))\\
)\\
(:action stack\\
  :parameters  (?ob ?underob)\\
  :precondition (and (clear ?underob) (holding ?ob))\\
  :effect (and (arm-empty) (clear ?ob) (on ?ob ?underob) (not (clear ?underob)) (not (holding ?ob)))\\
)\\
(:action unstack\\
  :parameters  (?ob ?underob)\\
  :precondition (and (on ?ob ?underob) (clear ?ob) (arm-empty))\\
  :effect (and (holding ?ob) (clear ?underob) (not (on ?ob ?underob)) (not (clear ?ob)) (not (arm-empty))))\\
)\\
}
Problem file:\\
\texttt{
(define (problem threeblocks)\\
(:domain blocksworld)\\
(:objects b1 b2 b3)\\
(:init \\
(arm-empty)\\
(on-table b1)\\
(on b2 b3)\\
(on-table b3)\\
(clear b1)\\
(clear b2)\\
)\\
(:goal\\
(and \\
(on b2 b1)\\
(on b1 b3)\\
(on-table b3))\\
)\\
}

\textbf{Answer:}

Action Sequence:\\
We need to build the blocks from bottom to top.\\
Third goal: b3 on table\\
Already satisfied in initial configuration\\
Second goal: b1 on b3\\
\texttt{(unstack b2 b3)\\
(putdown b2)\\
(pickup b1)\\
(stack b1 b3)\\}
First goal: b2 on b1\\
\texttt{
(pickup b2)\\
(stack b2 b1)}
\end{tcolorbox}

\subsection{Self-Validator}
\label{sec.validator_example}

\paragraph{Cooking:} 
For the Cooking planning domain, we use a total of four few-shot examples, which are given as follows.

\begin{tcolorbox}[title={Example 1},breakable]
\textbf{Question:}

Initial state: \\
\texttt{
(arm-empty)\\
(pot-empty pot1)\\
(pot-empty pot2)\\
(pot-empty pot3)\\}
Goal state:\\
\texttt{
(contain pot1 ingredient1)\\
(contain pot1 ingredient4)\\
(contain pot2 ingredient2)\\
(contain pot2 ingredient3)\\
(contain pot2 ingredient5)\\
(contain pot2 ingredient6)\\
(contain pot3 ingredient1)\\
(contain pot3 ingredient4)\\
(contain pot3 ingredient5)\\}
Examined action sequence:\\
\texttt{
(pick ingredient1)\\
(add ingredient1 pot1)\\
(add ingredient1 pot3)\\
(putdown ingredient1)\\
(pick ingredient2)\\
(add ingredient1 pot2)\\
(putdown ingredient2)\\
(pick ingredient3)\\
(add ingredient3 pot2)\\
(putdown ingredient3)\\
(pick ingredient4)\\
(add ingredient4 pot1)\\
(add ingredient4 pot3)\\
(putdown ingredient4)\\
(pick ingredient5)\\
(add ingredient5 pot2)\\
(add ingredient5 pot3)\\
(putdown ingredient5)\\
(pick ingredient6)\\
(add ingredient6 pot2)\\
(putdown ingredient6)\\
}

\textbf{Answer:}

Analysis:\\
Initial state:\\
pot1: empty\\
pot2: empty\\
pot3: empty\\
\texttt{(pick ingredient1)} result: ingredient1 in hand\\
\texttt{(add ingredient1 pot1)} result: \\
pot1: ingredient1\\
pot2: empty\\
pot3: empty\\
\texttt{(add ingredient1 pot3)} result: \\
pot1: ingredient1\\
pot2: empty\\
pot3: ingredient1 \\
\texttt{(putdown ingredient1)} result: hand empty\\
ingredient has been picked: ingredient1\\
\texttt{(pick ingredient2)} result: ingredient2 in hand\\
\texttt{(add ingredient1 pot2)} result:\\
action is wrong since currently ingredient2 is in hand instead of ingredient1\\
Final answer:\\
No, the action sequence is wrong, it cannot accomplish the goal.
\end{tcolorbox}

\begin{tcolorbox}[title={Example 2},breakable]
\textbf{Question:}

Initial state: \\
\texttt{
(arm-empty)\\
(pot-empty pot1)\\
(pot-empty pot2)\\\\
(pot-empty pot3)\\
}
Goal state:\\
\texttt{
(contain pot1 ingredient1)\\
(contain pot1 ingredient2)\\
(contain pot1 ingredient3)\\
(contain pot1 ingredient6)\\
(contain pot2 ingredient2)\\
(contain pot2 ingredient4)\\
(contain pot3 ingredient2)\\
(contain pot3 ingredient3)\\
(contain pot3 ingredient5)\\
}
Examined action sequence:\\
\texttt{
(pick ingredient1)\\
(add ingredient1 pot1)\\
(putdown ingredient1)\\
(pick ingredient2)\\
(add ingredient2 pot1)\\
(add ingredient2 pot2)\\
(add ingredient2 pot3)\\
(putdown ingredient2)\\
(pick ingredient2)\\
(add ingredient2 pot1)\\
(add ingredient2 pot3)\\
(putdown ingredient2)\\
(pick ingredient4)\\
(add ingredient4 pot2)\\
(putdown ingredient4)\\
(pick ingredient6)\\
(add ingredient6 pot1)\\
(putdown ingredient6)\\
}

\textbf{Answer:}

Analysis:\\
Initial state: \\
pot1: empty\\
pot2: empty\\
pot3: empty\\
\texttt{(pick ingredient1)} result: ingredient1 in hand\\
\texttt{(add ingredient1 pot1)} result: \\
pot1: ingredient1\\
pot2: empty\\
pot3: empty\\
\texttt{(putdown ingredient1)} result: hand empty\\
ingredient has been picked: ingredient1\\
\texttt{(pick ingredient2)} result: ingredient2 in hand\\
\texttt{(add ingredient2 pot1)} result:\\
pot1: ingredient1, ingredient2\\
pot2: empty\\
pot3: empty\\
\texttt{(add ingredient2 pot2)} result:\\
pot1: ingredient1, ingredient2\\
pot2: ingredient2\\
pot3: empty\\
\texttt{(add ingredient2 pot3)} result:\\
pot1: ingredient1, ingredient2\\
pot2: ingredient2\\
pot3: ingredient2\\
\texttt{(putdown ingredient2)} result: hand empty\\
ingredient has been picked: ingredient1, ingredient2\\
\texttt{(pick ingredient2)} result:\\
action is wrong since ingredient2 has already been picked \\
Final answer:\\
No, the action sequence is wrong, it cannot accomplish the goal.
\end{tcolorbox}

\begin{tcolorbox}[title={Example 3},breakable]
\textbf{Question:}

Initial state: \\
\texttt{
(arm-empty)\\
(pot-empty pot1)\\
(pot-empty pot2)\\
(pot-empty pot3)\\}
Goal state:\\
\texttt{
(contain pot1 ingredient2)\\
(contain pot1 ingredient5)\\
(contain pot2 ingredient2)\\
(contain pot2 ingredient4)\\
(contain pot2 ingredient5)\\
(contain pot2 ingredient6)\\
(contain pot3 ingredient1)\\
(contain pot3 ingredient5)\\}
Examined action sequence:\\
\texttt{
(pick ingredient2)\\
(add ingredient2 pot1)\\
(add ingredient2 pot2)\\
(putdown ingredient2)\\
(pick ingredient4)\\
(add ingredient4 pot2)\\
(putdown ingredient4)\\
(pick ingredient5)\\
(add ingredient5 pot1)\\
(add ingredient5 pot2)\\
(add ingredient5 pot3)\\
(putdown ingredient5)\\
(pick ingredient6)\\
(add ingredient6 pot2)\\
(putdown ingredient6)\\
}

\textbf{Answer:}

Analysis:\\
Initial state: \\
pot1: empty\\
pot2: empty\\
pot3: empty\\
\texttt{(pick ingredient2)} result: ingredient2 in hand\\
\texttt{(add ingredient2 pot1)} result: \\
pot1: ingredient2\\
pot2: empty\\
pot3: empty\\
\texttt{(add ingredient2 pot2)} result: \\
pot1: ingredient2\\
pot2: ingredient2\\
pot3: empty\\
\texttt{(putdown ingredient2)} result: hand empty\\
ingredient has been picked: ingredient2\\
\texttt{(pick ingredient4)} result: ingredient4 in hand\\
\texttt{(add ingredient4 pot2)} result: \\
pot1: ingredient2\\
pot2: ingredient2, ingredient4\\
pot3: empty\\
\texttt{(putdown ingredient4)} result: hand empty\\
ingredient has been picked: ingredient2, ingredient4\\
\texttt{(pick ingredient5)} result: ingredient5 in hand\\
\texttt{(add ingredient5 pot1)} result: \\
pot1: ingredient2, ingredient5\\
pot2: ingredient2, ingredient4\\
pot3: empty\\
\texttt{(add ingredient5 pot2)} result: \\
pot1: ingredient2, ingredient5\\
pot2: ingredient2, ingredient4, ingredient5\\
pot3: empty\\
\texttt{(add ingredient5 pot3)} result: \\
pot1: ingredient2, ingredient5\\
pot2: ingredient2, ingredient4, ingredient5\\
pot3: ingredient5\\
\texttt{(putdown ingredient5)} result: hand empty\\
ingredient has been picked: ingredient2, ingredient4, ingredient5\\
\texttt{(pick ingredient6)} result: ingredient6 in hand\\
\texttt{(add ingredient6 pot2)} result: \\
pot1: ingredient2, ingredient5\\
pot2: ingredient2, ingredient4, ingredient5, ingredient6\\
pot3: ingredient5\\
\texttt{(putdown ingredient6)} result: hand empty\\
ingredient has been picked: ingredient2, ingredient4, ingredient5, ingredient6\\
No error found in action sequence.\\
Goal state check: \\
final state: \\
pot1: ingredient2, ingredient5\\
pot2: ingredient2, ingredient4, ingredient5, ingredient6\\
pot3: ingredient5\\
goal state: \\
\texttt{(contain pot1 ingredient2)\\
(contain pot1 ingredient5)\\
(contain pot2 ingredient2)\\
(contain pot2 ingredient4)\\
(contain pot2 ingredient5)\\
(contain pot2 ingredient6)\\
(contain pot3 ingredient1)\\
(contain pot3 ingredient5)\\}
The final state is not the same as the goal state. pot3 should contain ingredient1. Goal is not accomplished.\\
Final answer:\\
No, the action sequence is wrong, it cannot accomplish the goal. pot3 should contain ingredient1.
\end{tcolorbox}

\begin{tcolorbox}[title={Example 4},breakable]
\textbf{Question:}

Initial state: \\
\texttt{
(arm-empty)\\
(pot-empty pot1)\\
(pot-empty pot2)\\
(pot-empty pot3)\\}
Goal state:\\
\texttt{
(contain pot1 ingredient2)\\
(contain pot1 ingredient3)\\
(contain pot1 ingredient4)\\
(contain pot1 ingredient6)\\
(contain pot2 ingredient3)\\
(contain pot2 ingredient5)\\
(contain pot3 ingredient4)\\
(contain pot3 ingredient5)\\}
Examined action sequence:\\
\texttt{
(pick ingredient2)\\
(add ingredient2 pot1)\\
(putdown ingredient2)\\
(pick ingredient3)\\
(add ingredient3 pot1)\\
(add ingredient3 pot2)\\
(putdown ingredient3)\\
(pick ingredient4)\\
(add ingredient4 pot1)\\
(add ingredient4 pot3)\\
(putdown ingredient4)\\
(pick ingredient5)\\
(add ingredient5 pot2)\\
(add ingredient5 pot3)\\
(putdown ingredient5)\\
(pick ingredient6)\\
(add ingredient6 pot1)\\
(putdown ingredient6)\\
}

\textbf{Answer:}

Analysis:\\
Initial state: \\
pot1: empty\\
pot2: empty\\
pot3: empty\\
\texttt{(pick ingredient2)} result: ingredient2 in hand\\
\texttt{(add ingredient2 pot1)} result: \\
pot1: ingredient2\\
pot2: empty\\
pot3: empty\\
\texttt{(putdown ingredient2)} result: hand empty\\
ingredient has been picked: ingredient2\\
\texttt{(pick ingredient3)} result: ingredient3 in hand\\
\texttt{(add ingredient3 pot1)} result: \\
pot1: ingredient2, ingredient3\\
pot2: empty\\
pot3: empty\\
\texttt{(add ingredient3 pot2)} result: \\
pot1: ingredient2, ingredient3\\
pot2: ingredient3\\
pot3: empty\\
\texttt{(putdown ingredient3)} result: hand empty\\
ingredient has been picked: ingredient2, ingredient3\\
\texttt{(pick ingredient4)} result: ingredient4 in hand\\
\texttt{(add ingredient4 pot1)} result: \\
pot1: ingredient2, ingredient3, ingredient4\\
pot2: ingredient3\\
pot3: empty\\
\texttt{(add ingredient4 pot3)} result: \\
pot1: ingredient2, ingredient3, ingredient4\\
pot2: ingredient3\\
pot3: ingredient4\\
\texttt{(putdown ingredient4)} result: hand empty\\
ingredient has been picked: ingredient2, ingredient3, ingredient4\\
\texttt{(pick ingredient5)} result: ingredient5 in hand\\
\texttt{(add ingredient5 pot2)} result: \\
pot1: ingredient2, ingredient3, ingredient4\\
pot2: ingredient3, ingredient5\\
pot3: ingredient4\\
\texttt{(add ingredient5 pot3)} result: \\
pot1: ingredient2, ingredient3, ingredient4\\
pot2: ingredient3, ingredient5\\
pot3: ingredient4, ingredient5\\
\texttt{(putdown ingredient5)} result: hand empty\\
ingredient has been picked: ingredient2, ingredient3, ingredient4, ingredient5\\
\texttt{(pick ingredient6)} result: ingredient6 in hand\\
\texttt{(add ingredient6 pot1)} result: \\
pot1: ingredient2, ingredient3, ingredient4, ingredient6\\
pot2: ingredient3, ingredient5\\
pot3: ingredient4, ingredient5\\
\texttt{(putdown ingredient6)} result: hand empty\\
ingredient has been picked: ingredient2, ingredient3, ingredient4, ingredient5, ingredient6\\
No error found in action sequence.\\
Goal state check: \\
final state: \\
pot1: ingredient2, ingredient3, ingredient4, ingredient6\\
pot2: ingredient3, ingredient5\\
pot3: ingredient4, ingredient5\\
goal state: \\
\texttt{
(contain pot1 ingredient2)\\
(contain pot1 ingredient3)\\
(contain pot1 ingredient4)\\
(contain pot1 ingredient6)\\
(contain pot2 ingredient3)\\
(contain pot2 ingredient5)\\
(contain pot3 ingredient4)\\
(contain pot3 ingredient5)\\}
The final state is the same as the goal state. Goal is accomplished.\\
Final answer:\\
Yes, the action sequence is correct, it can accomplish the task.
\end{tcolorbox}

\paragraph{Ball Moving:} 
For the Ball Moving planning domain, we use a total of five few-shot examples, which are given as follows.

\begin{tcolorbox}[title={Example 1},breakable]
\textbf{Question:}

Robot and ball initial state: \\
\texttt{
(robot-at robot1 room1)\\
(at ball1 room4)\\
(at ball2 room3)\\
(at ball3 room4)\\}
Goal state:\\
\texttt{
(at ball1 room4)\\
(at ball2 room4)\\
(at ball3 room3)\\}
Examined action sequence:\\
\texttt{
(move robot1 room1 room3)\\
(pick ball2 room3)\\
(move robot1 room3 room4)\\
(drop ball2 room3)\\
(pick ball1 room4)\\
(move robot1 room4 room3)\\
(drop ball1 room3)\\
(pick ball3 room4)\\
(move robot1 room3 room4)\\
(drop ball3 room3)\\
}

\textbf{Answer:}

Analysis:\\
Initial state: \texttt{(robot-at robot1 room1), (at ball1 room4), (at ball2 room3), (at ball3 room4)}\\
\texttt{(move robot1 room1 room3)} \\
precondition: \texttt{(robot-at robot1 room1)}\\
current state: \texttt{(robot-at robot1 room1)}\\
current state is the same as the precondition, action is correct\\
change state to: \texttt{(robot-at robot1 room3), (at ball1 room4), (at ball2 room3), (at ball3 room4)\\}
\texttt{(pick ball2 room3)} \\
precondition 1: \texttt{(at ball2 room3)}\\
current state: \texttt{(at ball2 room3)}\\
current state is the same as the precondition 1\\
precondition 2: robot1 and ball2 in the same room\\
current state: \texttt{(robot-at robot1 room3), (at ball2 room3)}\\
robot1 is in the same room as ball2\\
two preconditions are correct, action is correct\\
change state to: \texttt{(robot-at robot1 room3), (at ball1 room4), (at ball2 hand), (at ball3 room4)}\\
\texttt{(move robot1 room3 room4)} \\
precondition: \texttt{(robot-at robot1 room3)}\\
current state: \texttt{(robot-at robot1 room3)}\\
current state is the same as the precondition, action is correct\\
change state to: \texttt{(robot-at robot1 room4), (at ball1 room4), (at ball2 hand), (at ball3 room4)\\}
\texttt{(drop ball2 room3)} \\
precondition: \texttt{(robot-at robot1 room3)}\\
current state: \texttt{(robot-at robot1 room4)}\\
the current state is not the same as the precondition, action is wrong\\
analysis stops due to error\\
Final answer:\\
No, the action sequence is wrong, it cannot accomplish the goal.
\end{tcolorbox}

\begin{tcolorbox}[title={Example 2},breakable]
\textbf{Question:}

Robot and ball initial state: \\
\texttt{
(robot-at robot1 room1)\\
(at ball1 room3)\\
(at ball2 room4)\\
(at ball3 room2)\\}
Goal state:\\
\texttt{
(at ball1 room2)\\
(at ball2 room3)\\
(at ball3 room1)\\}
Examined action sequence:\\
\texttt{
(pick ball2 room4)\\
(move robot1 room2 room3)\\
(drop ball2 room3)\\
(move robot1 room3 room2)\\
(pick ball1 room3)\\
(move robot1 room2 room1)\\
(drop ball1 room2)\\
(move robot1 room1 room2)\\
(pick ball3 room2)\\
(move robot1 room2 room1)\\
(drop ball3 room1)\\
}

\textbf{Answer:}

Analysis:\\
Initial state: \texttt{(robot-at robot1 room1), (at ball1 room3), (at ball2 room4), (at ball3 room2)\\}
\texttt{(pick ball2 room4)} \\
precondition 1: \texttt{(at ball2 room4)}\\
current state: \texttt{(at ball2 room4)}\\
current state is the same as the precondition 1\\
precondition 2: robot1 and ball2 in the same room\\
current state: \texttt{(robot-at robot1 room1), (at ball2 room4)\\}
robot1 is not in the same room as ball2\\
preconditions are not fullfilled, action is wrong\\
analysis stops due to error\\
Final answer:\\
No, the action sequence is wrong, it cannot accomplish the goal.
\end{tcolorbox}

\begin{tcolorbox}[title={Example 3},breakable]
\textbf{Question:}

Robot and ball initial state: \\
\texttt{(robot-at robot1 room2)\\
(at ball1 room2)\\
(at ball2 room1)\\
(at ball3 room3)\\}
Goal state:\\
\texttt{(at ball1 room1)\\
(at ball2 room1)\\
(at ball3 room3)\\}
Examined action sequence:\\
\texttt{
(move robot1 room2 room1)\\
(pick ball1 room1)\\
(move robot1 room1 room2)\\
(drop ball1 room2)\\
(pick ball2 room1)\\
(move robot1 room2 room1)\\
(drop ball2 room1)\\
}

\textbf{Answer:}

Analysis:\\
Initial state: \texttt{(robot-at robot1 room2), (at ball1 room2), (at ball2 room1), (at ball3 room3)\\}
\texttt{(move robot1 room2 room1)} \\
precondition: \texttt{(robot-at robot1 room2)}\\
current state: \texttt{(robot-at robot1 room2)}\\
current state is the same as the precondition, action is correct\\
change state to: \texttt{(robot-at robot1 room1), (at ball1 room2), (at ball2 room1), (at ball3 room3)\\}
\texttt{(pick ball1 room1)} \\
precondition 1: \texttt{(at ball1 room1)}\\
current state: \texttt{(at ball1 room2)}\\
current state is not the same as the precondition 1\\
preconditions are not fullfilled, action is wrong\\
analysis stops due to error\\
Final answer:\\
No, the action sequence is wrong, it cannot accomplish the goal.
\end{tcolorbox}

\begin{tcolorbox}[title={Example 4},breakable]
\textbf{Question:}

Robot and ball initial state: \\
\texttt{
(robot-at robot1 room4)\\
(at ball1 room2)\\
(at ball2 room4)\\
(at ball3 room3)\\}
Goal state:\\
\texttt{
(at ball1 room3)\\
(at ball2 room4)\\
(at ball3 room4)\\}
Examined action sequence:\\
\texttt{
(move robot1 room4 room2)\\
(pick ball1 room2)\\
(move robot1 room2 room3)\\
(drop ball1 room3)\\}

\textbf{Answer:}

Analysis:\\
Initial state: \texttt{(robot-at robot1 room4), (at ball1 room2), (at ball2 room4), (at ball3 room3)\\}
\texttt{(move robot1 room4 room2)} \\
precondition: \texttt{(robot-at robot1 room4)}\\
current state: \texttt{(robot-at robot1 room4)}\\
current state is the same as the precondition, action is correct\\
change state to: \texttt{(robot-at robot1 room2), (at ball1 room2), (at ball2 room4), (at ball3 room3)\\}
\texttt{(pick ball1 room2)} \\
precondition 1: \texttt{(at ball1 room2)}\\
current state: \texttt{(at ball1 room2)}\\
current state is the same as the precondition 1\\
precondition 2: robot1 and ball1 in the same room\\
current state: \texttt{(robot-at robot1 room2), (at ball1 room2)\\}
robot1 is in the same room as ball1\\
two preconditions are correct, action is correct\\
change state to: \texttt{(robot-at robot1 room2), (at ball1 hand), (at ball2 room4), (at ball3 room3)\\}
\texttt{(move robot1 room2 room3)} \\
precondition: \texttt{(robot-at robot1 room2)}\\
current state: \texttt{(robot-at robot1 room2)}\\
current state is the same as the precondition, action is correct\\
change state to: \texttt{(robot-at robot1 room3), (at ball1 hand), (at ball2 room4), (at ball3 room3)\\}
\texttt{(drop ball1 room3)} \\
precondition: \texttt{(robot-at robot1 room3)}\\
current state: \texttt{(robot-at robot1 room3)}\\
current state is the same as the precondition, action is correct\\
change state to: \texttt{(robot-at robot1 room3), (at ball1 room3), (at ball2 room4), (at ball3 room3)\\}
No error found in action sequence.\\
Goal state check: \\
final state: \texttt{(at ball1 room3), (at ball2 room4), (at ball3 room3)\\}
goal state: \texttt{(at ball1 room3), (at ball2 room4), (at ball3 room4)\\}
The final state is not the same as the goal state. The goal is not accomplished.\\
Final answer:\\
No, the action sequence is wrong, it cannot accomplish the goal.
\end{tcolorbox}

\begin{tcolorbox}[title={Example 5},breakable]
\textbf{Question:}

Robot and ball initial state: \\
\texttt{
(robot-at robot1 room1)\\
(at ball1 room4)\\
(at ball2 room1)\\
(at ball3 room1)\\}
Goal state:\\
\texttt{
(at ball1 room2)\\
(at ball2 room1)\\
(at ball3 room3)\\}
Examined action sequence:\\
\texttt{
(pick ball3 room1)\\
(move robot1 room1 room3) \\
(drop ball3 room3)\\
(move robot1 room3 room4) \\
(pick ball1 room4) \\
(move robot1 room4 room2)\\
(drop ball1 room2)\\
}

\textbf{Answer:}

Analysis:\\
Initial state: \texttt{(robot-at robot1 room1), (at ball1 room4), (at ball2 room1), (at ball3 room1)\\}
\texttt{(pick ball3 room1)} \\
precondition 1: \texttt{(at ball3 room1)}\\
current state: \texttt{(at ball3 room1)}\\
current state is the same as the precondition 1\\
precondition 2: robot1 and ball3 in the same room\\
current state: \texttt{(robot-at robot1 room1), (at ball3 room1)\\}
robot1 is in the same room as ball1\\
two preconditions are correct, action is correct\\
change state to: \texttt{(robot-at robot1 room1), (at ball1 room4), (at ball2 room1), (at ball3 hand)\\}
\texttt{(move robot1 room1 room3)} \\
precondition: \texttt{(robot-at robot1 room1)\\}
current state: \texttt{(robot-at robot1 room1)\\}
current state is the same as the precondition, action is correct\\
change state to: \texttt{(robot-at robot1 room3), (at ball1 room4), (at ball2 room1), (at ball3 hand)\\}
\texttt{(drop ball3 room3)} \\
precondition: \texttt{(robot-at robot1 room3)}\\
current state: \texttt{(robot-at robot1 room3)}\\
current state is the same as the precondition, action is correct\\
change state to: \texttt{(robot-at robot1 room3), (at ball1 room4), (at ball2 room1), (at ball3 room3)\\}
\texttt{(move robot1 room3 room4)} \\
precondition: \texttt{(robot-at robot1 room3)}\\
current state: \texttt{(robot-at robot1 room3)}\\
current state is the same as the precondition, action is correct\\
change state to: \texttt{(robot-at robot1 room4), (at ball1 room4), (at ball2 room1), (at ball3 room3)\\}
\texttt{(pick ball1 room4)} \\
precondition 1: \texttt{(at ball1 room4)}\\
current state: \texttt{(at ball1 room4)}\\
current state is the same as the precondition 1\\
precondition 2: robot1 and ball1 in the same room\\
current state: \texttt{(robot-at robot1 room4), (at ball1 room4)}\\
robot1 is in the same room as ball1\\
two preconditions are correct, action is correct\\
change state to: \texttt{(robot-at robot1 room4), (at ball1 hand), (at ball2 room1), (at ball3 room3)\\}
\texttt{(move robot1 room4 room2)} \\
precondition: \texttt{(robot-at robot1 room4)}\\
current state: \texttt{(robot-at robot1 room4)}\\
current state is the same as the precondition, action is correct\\
change state to:  \texttt{(robot-at robot1 room2), (at ball1 hand), (at ball2 room1), (at ball3 room3)\\}
\texttt{(drop ball1 room2)} \\
precondition: \texttt{(robot-at robot1 room2)}\\
current state: \texttt{(robot-at robot1 room2)}\\
current state is the same as the precondition, action is correct\\
change state to:  \texttt{(robot-at robot1 room2), (at ball1 room2), (at ball2 room1), (at ball3 room3)\\}
No error found in action sequence.\\
Goal state check: \\
final state: \texttt{(at ball1 room2), (at ball2 room1), (at ball3 room3)}\\
goal state: \texttt{(at ball1 room2), (at ball2 room1), (at ball3 room3)}\\
The final state is the same as the goal state. The goal is accomplished.\\
Final answer:\\
Yes, the action sequence is correct, it can accomplish the task.
\end{tcolorbox}

\paragraph{Blocksworld:} 
For the Blocksworld planning domain, we use a total of six few-shot examples, which are given as follows.

\begin{tcolorbox}[title={Example 1},breakable]
\textbf{Question:}

Block initial state: \\
\texttt{
(on-table b1)\\
(on b2 b1)\\
(on b3 b2)\\}
Goal state:\\
\texttt{
(on b1 b2)\\
(on b2 b3)\\
(on-table b3)\\}
Examined action sequence:\\
\texttt{
(unstack b3 b2)\\
(putdown b3)\\
(unstack b2 b1)\\
(putdown b2)\\
(unstack b1 b2)\\
(putdown b1)\\
}

\textbf{Answer:}

Analysis:\\
Initial: b1 on the table, b2 on b1, b3 on b2\\
\texttt{(unstack b3 b2)} result: b1 on the table, b2 on b1, b3 in the hand\\
\texttt{(putdown b3)} result: b1 on the table, b2 on b1, b3 on the table\\
\texttt{(unstack b2 b1)} result: b1 on the table, b2 in hand, b3 on the table\\
\texttt{(putdown b2)} result: b1 on the table, b2 on the table, b3 on the table\\
\texttt{(unstack b1 b2)} result: the action is wrong since b1 is not on top of b2\\
analysis stops due to error\\
Final answer:\\
No, the action sequence is wrong, it cannot accomplish the goal.
\end{tcolorbox}

\begin{tcolorbox}[title={Example 2},breakable]
\textbf{Question:}

Block initial state: \\
\texttt{
(on b1 b3)\\
(on-table b2)\\
(on-table b3)\\}
Goal state:\\
\texttt{
(on b3 b2)\\
(on b2 b1)\\
(on-table b1)\\}
Examined action sequence:\\
\texttt{
(unstack b1 b3)\\
(putdown b1)\\
(pickup b2)\\
(stack b2 b1)\\
(pickup b3)\\
(stack b3 b2)\\
}

\textbf{Answer:}

Analysis:\\
Initial: b1 on b3, b2 on the table, b3 on the table\\
\texttt{(unstack b1 b3)} result: b1 in hand, b2 on the table, b3 on the table\\
\texttt{(putdown b1)} result: b1 on the table, b2 on the table, b3 on the table\\
\texttt{(pickup b2)} result: b1 on the table, b2 in hand, b3 on the table\\
\texttt{(stack b2 b1)} result: b1 on the table, b2 on b1, b3 on the table\\
\texttt{(pickup b3)} result: b1 on the table, b2 on b1, b3 in hand\\
\texttt{(stack b3 b2)} result: b1 on the table, b2 on b1, b3 on b2\\
No error found in action sequence.\\
Goal state check: the final state is the same as the goal state\\
Final answer:\\
Yes, the action sequence is correct, it can accomplish the task.
\end{tcolorbox}

\begin{tcolorbox}[title={Example 3},breakable]
\textbf{Question:}

Block initial state: \\
\texttt{
(on b1 b3)\\
(on-table b2)\\
(on b3 b2)\\}
Goal state:\\
\texttt{
(on b2 b3)\\
(on b3 b1)\\
(on-table b1)\\}
Examined action sequence:\\
\texttt{
(unstack b1 b3)\\
(putdown b1)\\
(unstack b3 b2)\\
(putdown b3)\\
(pickup b2)\\
(stack b2 b3)\\
}

\textbf{Answer:}

Analysis:\\
Initial: b1 on b3, b2 on the table, b3 on b2\\
\texttt{(unstack b1 b3)} result: b1 in hand, b2 on the table, b3 on b2\\
\texttt{(putdown b1)} result: b1 on the table, b2 on the table, b3 on b2\\
\texttt{(unstack b3 b2)} result: b1 on the table, b2 on the table, b3 in hand\\
\texttt{(putdown b3)} result: b1 on the table, b2 on the table, b3 on the table\\
\texttt{(pickup b2)} result: b1 on the table, b2 in hand, b3 on the table\\
\texttt{(stack b2 b3)} result: b1 on the table, b2 on b3, b3 on the table\\
No error found in action sequence.\\
Goal state check: the final state is not the same as the goal state\\
Final answer:\\
No, the action sequence is wrong, it cannot accomplish the goal.
\end{tcolorbox}

\begin{tcolorbox}[title={Example 4},breakable]
\textbf{Question:}

Block initial state: \\
\texttt{
(on b1 b3)\\
(on-table b2)\\
(on-table b3)\\}
Goal state:\\
\texttt{
(on b1 b3)\\
(on b3 b2)\\
(on-table b2)\\}
Examined action sequence:\\
\texttt{
(pickup b2)\\
(stack b2 b1)\\
(pickup b3)\\
(stack b3 b2)\\
(pickup b1)\\
(stack b1 b3)\\
}

\textbf{Answer:}

Analysis:\\
Initial: b1 on b3, b2 on the table, b3 on the table\\
\texttt{(pickup b2)} result: b1 on b3, b2 in hand, b3 on the table\\
\texttt{(stack b2 b1)} result: b1 on b3, b2 on b1, b3 on the table\\
\texttt{(pickup b3)} result: the action is wrong since b3 is not clear to move\\
analysis stops due to error\\
Final answer:\\
No, the action sequence is wrong, it cannot accomplish the goal.
\end{tcolorbox}

\begin{tcolorbox}[title={Example 5},breakable]
\textbf{Question:}

Block initial state: \\
\texttt{
(on-table b1)\\
(on-table b2)\\
(on-table b3)\\}
Goal state:\\
\texttt{
(on b2 b3)\\
(on b3 b1)\\
(on-table b1)\\}
Examined action sequence:\\
\texttt{
(pickup b3)\\
(stack b3 b1)\\
(pickup b2)\\
(stack b2 b3)\\
}

\textbf{Answer:}

Analysis:\\
Initial: b1 on the table, b2 on the table, b3 on the table\\
\texttt{(pickup b3)} result: b1 on the table, b2 on the table, b3 in hand\\
\texttt{(stack b3 b1)} result: b1 on the table, b2 on the table, b3 on b1\\
\texttt{(pickup b2)} result: b1 on the table, b2 in hand, b3 on b1\\
\texttt{(stack b2 b3)} result: b1 on the table, b2 on b3, b3 on b1\\
No error found in action sequence.\\
Goal state check: the final state is the same as the goal state\\
Final answer:\\
Yes, the action sequence is correct, it can accomplish the task.
\end{tcolorbox}

\begin{tcolorbox}[title={Example 6},breakable]
\textbf{Question:}

Block initial state: \\
\texttt{
(on b1 b2)\\
(on b2 b3)\\
(on-table b3)\\}
Goal state:\\
\texttt{
(on b1 b3)\\
(on b3 b2)\\
(on-table b2)\\}
Examined action sequence:\\
\texttt{
(unstack b1 b2)\\
(putdown b1)\\
(pickup b1)\\
(stack b1 b3)\\
(unstack b3 b2)\\
(putdown b3)\\
(pickup b3)\\
(stack b3 b2)\\
(unstack b2 b3)\\
(putdown b2)\\
(pickup b2)\\
(stack b2 b3)\\
}

\textbf{Answer:}

Analysis:\\
Initial: b1 on b2, b2 on b3, b3 on the table\\
\texttt{(unstack b1 b2)} result: b1 in hand, b2 on b3, b3 on the table\\
\texttt{(putdown b1)} result: b1 on the table, b2 on b3, b3 on the table\\
\texttt{(pickup b1)} result: b1 in hand, b2 on b3, b3 on the table\\
\texttt{(stack b1 b3)} result: the action is wrong since b3 is not clear to move\\
analysis stops due to error\\
Final answer:\\
No, the action sequence is wrong, it cannot accomplish the goal.
\end{tcolorbox}







\end{document}